\documentclass[11pt,oneside,reqno]{amsart}

\usepackage{amssymb}
\usepackage{algpseudocode}
\usepackage{algorithm}
\usepackage{verbatim}
\usepackage{graphicx}
\usepackage{placeins}
\usepackage{listings}
\usepackage{xcolor}
\usepackage{subcaption}
\usepackage[noadjust]{cite}
\RequirePackage{dsfont} 
 \scrollmode
\allowdisplaybreaks
\usepackage{graphicx}
\usepackage{epstopdf}
\usepackage{hyperref} 
\usepackage{mathtools}
\usepackage{float}

\usepackage{wrapfig}

\usepackage{amsmath}
\usepackage{tikz}
\usepackage{enumerate}
\usepackage{xcolor}


\definecolor{codegreen}{rgb}{0,0.6,0}
\definecolor{codegray}{rgb}{0.5,0.5,0.5}
\definecolor{codepurple}{rgb}{0.58,0,0.82}
\definecolor{backcolour}{rgb}{0.95,0.95,0.92}

\lstdefinestyle{list_style}{
  backgroundcolor=\color{backcolour}, commentstyle=\color{codegreen},
  keywordstyle=\color{magenta},
  numberstyle=\tiny\color{codegray},
  stringstyle=\color{codepurple},
  basicstyle=\ttfamily\footnotesize,
  breakatwhitespace=false,         
  breaklines=true,                 
  captionpos=b,                    
  keepspaces=true,                 
  numbers=left,                    
  numbersep=5pt,                  
  showspaces=false,                
  showstringspaces=false,
  showtabs=false,                  
  tabsize=2
}

\newcommand{\xdasharrow}[2][->]{
\tikz[baseline=-\the\dimexpr\fontdimen22\textfont2\relax]{
\node[anchor=south,font=\scriptsize, inner ysep=1.5pt,outer xsep=2.2pt](x){#2};
\draw[shorten <=3.4pt,shorten >=3.4pt,dashed,#1](x.south west)--(x.south east);
}
}

\newcommand{\DEBUG}{}

\ifdefined\DEBUG%

  \def\rem#1{{\marginpar{\raggedright\scriptsize #1}}}
  \newcommand{\pmr}[1]{\rem{\color{blue}{$\bullet$ #1}}}
  \newcommand{\ppr}[1]{\rem{\color{red}{$\bullet$ #1}}}
 \else%

  \newcommand{\ppr}[1]{}
  \newcommand{\pmr}[1]{}
 \fi

\def\rho{\varrho_1}

\def\rd{\,{\mathrm d}}

\newcommand{\norm}[1]{\left\lVert#1\right\rVert}

\theoremstyle{plain}
\newtheorem{theorem}{Theorem}
\newtheorem{lemma}{Lemma}
\newtheorem{fact}{Fact}

\DeclareMathOperator{\sign}{sign}

\theoremstyle{definition}
\newtheorem{remark}{Remark}

\begin{document}

\title[Deep learning-based estimation of time-dependent parameters]{Deep learning-based estimation of time-dependent parameters in Markov models with application to nonlinear regression and SDEs}
\author[A. Ka\l u\.za]{	Andrzej Ka\l u\.za} 
\address{AGH University of Krakow,
Faculty of Applied Mathematics,
 Al. A.~Mickiewicza 30, 30-059 Krak\'ow, Poland}
 \email{akaluza@agh.edu.pl}
\author[P. M. Morkisz]{Pawe{\l } M. Morkisz}
 \address{AGH University of Krakow,
 	Faculty of Applied Mathematics,
 	Al. A.~Mickiewicza 30, 30-059 Krak\'ow, Poland}
 \email{morkiszp@agh.edu.pl}
 
\author[B. Mulewicz]{Bart{\l }omiej Mulewicz}\address{Aigorithmics sp. z o. o., ul. Bronowicka 10A, 30-084 Krak\'ow}\email{bartekmulewicz@gmail.com}

\author[P. Przyby\l owicz]{Pawe{\l } Przyby\l owicz}
 \address{AGH University of Krakow,
Faculty of Applied Mathematics,
 Al. A.~Mickiewicza 30, 30-059 Krak\'ow, Poland and Enerco Sp. z o.o., ul. Gotarda 9, 02-683 Warsaw, Poland}
 \email{pprzybyl@agh.edu.pl, pp@enerco.pl, corresponding author}

\author[M. Wi\c{a}cek]{Martyna Wi\c{a}cek}
 \address{Aigorithmics sp. z o. o., ul. Bronowicka 10A, 30-084 Krak\'ow and AGH University of Krakow,
Faculty of Applied Mathematics,
 Al. A.~Mickiewicza 30, 30-059 Krak\'ow, Poland}
\email{martynawiacek@agh.edu.pl}

\begin{abstract}
We present a novel deep learning method for estimating time-dependent parameters in Markov processes through discrete sampling. Departing from conventional machine learning, our approach reframes parameter approximation as an optimization problem using the maximum likelihood approach. Experimental validation focuses on parameter estimation in multivariate regression and stochastic differential equations (SDEs). Theoretical results show that the real solution is close to SDE with parameters approximated using our neural network-derived under specific conditions. Our work contributes to SDE-based model parameter estimation, offering a versatile tool for diverse fields.

\end{abstract}
\maketitle
\section{Introduction}
In real-world scenarios, parameters in problems seldom exhibit constant values but rather evolve over time, posing a challenge for accurate estimation. This paper introduces a framework facilitating robust estimation of time-dependent parameters in practical applications, leveraging powerful open-source libraries. Our approach addresses the complexities associated with parameter estimation in the context of nonlinear regression and Stochastic Differential Equation (SDE)-based models, wherein the parameters are inherently time-varying. In this paper, we generalize those two cases into an abstract concept of Markov models and we formalize the approach on how to approximate time-dependent parameters of Markov models in general. Then, the results presented herein demonstrate the effectiveness of our proposed methodology in handling such dynamic parameter scenarios.

The problem of nonlinear regression was already addressed by authors of \cite{DSM}, we present its generalization to the case of multivariate regression. The nonlinear regression parameters estimation has applications in numerous real-world problems, e.g., in finances \cite{adamowski2012comparison,adrian2019nonlinearity,alfouhaili2020impact}, pharmacokinetics modeling \cite{saganuwan2021application}, and agriculture \cite{archontoulis2015nonlinear}.

Estimation of parameters in SDE-based models is a complex problem with important practical applications in many fields, for example, in finances, energy prices, or consumption forecasting. Estimating time-dependent parameters is a challenge due to the multitude of estimated values, and the most common approach is simplification using piecewise-constant functions. We propose a novel method for estimating time-dependent parameters that is based on neural networks and which extends the approach known from \cite{DSM} for the regression in the heteroscedasticity case.

The problem of approximating parameters of stochastic differential equations, known as calibration, has received extensive research attention. While various methods have been explored, the conventional statistical techniques such as maximum likelihood estimation (MLE), least squares methods (LSM), and the method of moments remain the most widely employed approaches. The first studies examining conditions and constraints for likelihood estimation of unknown parameters in diffusion processes date back to the late 1980s and early 1990s, as described in  \cite{florens1989approximate} and \cite{yoshida1992estimation}. Subsequent research expanded on the MLE-based approaches, e.g., in \cite{delattre2013maximum}, and \cite{pedersen1995new}. Further analyses for the equations involving random effects driven by fractional Brownian motion were presented in\cite{dai2021maximum}, and \cite{prakasa2023maximum}. Least squared estimators have also been explored in works such as \cite{long2017least} or \cite{zhen2023least}. Additionally, the method of moments was effectively employed in \cite{andersen1999efficient} for modeling ARCH, GARCH, and EGARCH models. The Generalized Method of Moments (GMM) developed by Hansen in 1982 \cite{hansen1982large} and its further extensions have played a significant role in the literature related to the parameter and state estimation problems in SDEs. An intriguing contribution to the generalized method of moments is documented in\cite{otunuga2019local}, where the concept of local lagged adapted generalized method of moments was introduced and described.


There are also studies dedicated to the estimation of parameters of specific models. For instance, \cite{tian2014estimation} focuses on mean-reverting stochastic systems, and \cite{feron2015calibration} on a two-factor model.

Furthermore, more advanced methods have been devised for calibrating parameters of SDEs. For instance, in the paper \cite{tian2012calibration}, Tian and Ge propose using Particle Swarm Optimization (PSO) algorithm for searching the parameter space, along with an analysis of simulated maximum likelihood, where the transition density is approximated with specific kernel density function. In another approach outlined in \cite{riseth2017operator} by Riseth and Taylor-King, they describe a method where  a finite-dimensional approximation of the Koopman operator is matched with the implied Koopman operator, generated by an extended dynamic mode decomposition approximation. Notably, this approach can lead to objective evaluation cost that is independent of the number of samples for many dynamical systems. Wei, in \cite{wei2019estimation}, analyzed state and parameter estimation for nonlinear stochastic systems using extended Kalman filtering. Additionally, intriguing findings are presented in \cite{liao2019learning}, where Logsig-RNN model, inspired by the numerical approximation theory of SDEs, is presented as an efficient and robust algorithm for learning a functional representation from streamed data. 

In most of these studies, authors develop theory for multidimensional SDEs but typically limit their practical applications to one-dimensional SDEs due to the computational challenges. Notably, in each of these cases, the parameters are assumed to be constant in time. The concept of time dimension reduction is discussed as a possibility only in \cite{liao2019learning} and \cite{riseth2017operator} (available as preprints). 

In this paper, we present a novel approach that introduces an estimation technique for unknown parameters based on a discrete sampling of Markov processes, particularly for multivariate regression or SDE-based models. Our algorithm centers on the definition of a suitable loss function rooted in the maximum likelihood framework, enabling us to transform our approximation task into an optimization problem. This, in turn, allows us to use deep learning techniques. It's worth highlighting that our approach is not black-box model. Instead, we strategically select a suitable stochastic difference equation model and employ neural network solely to calibrate its parameters based on a single trajectory. This characterizes our approach more of a grey-box model. All the codes used for the experiments reported in this paper can be found online: \url{https://github.com/mwiacek8/DL-based-estimation}.

This article is organized as follows. Section \ref{section2} lays the foundation by introducing the general setting for discrete-time Markov processes and the construction of the loss function. In Section~\ref{section3}, we provide the formula for the maximum likelihood of a multivariate regression and also construction, together with the results of our deep learning model. Section \ref{section4} is dedicated to the application of this approach for SDEs, accompanied by theoretical results and numerical experiments.
Section \ref{section5} discusses the results and the next steps. We also attach three appendices, in Appendix \ref{appendixA} we present the Python code utilized in our experiments, and Appendix \ref{appendixB} contains various proofs for the reader's convenience. Appendix \ref{appendixC}
contains procedure for Monte Carlo forecasting for SDEs. 
\section{General setting for discrete time Markov processes}\label{section2}
Aim of this paper is to convert the given problem in a form that neural networks are applicable. We provide a general setting based on discrete Markov processes that allows it. Next, we settle two problems in this setting: the problem of nonlinear regression with varying variance and time-dependent parameter estimation for SDEs.

Let
  $$x_1,x_2,\ldots,x_n$$ be observed realizations at the fixed time points $t_1<t_2<\ldots t_n$, $t_i\in [0,T]$, of the sequence of $\mathbb{R}^d$-valued random vectors $$X_1,X_2,\ldots,X_n,$$ for which we assume the (discrete) Markov property, i.e.:
    \begin{equation}
    \label{disc_markov_prop}
        \mathbb{P}(X_{k+1}\in A \ | \  X_1,\ldots,X_k)=\mathbb{P}(X_{k+1}\in A \ | \ X_k),
    \end{equation}
    for all Borel sets $A$ and $k=0,1,\ldots,n-1$, where $\mathbb{P}(X_{1}\in A \ | \ X_0):=\mathbb{P}(X_{1}\in A)$
We assume that for all $k=0,1,\ldots,n-1$ there exists conditional density $f_{X_{k+1}| X_{k}}(\ \cdot\ | x_k)$, with $f_{X_{1}| X_{0}}(\ \cdot\ | x_0):=f_{X_1}(\ \cdot\ )$. We assume that $f_{X_{k+1}| X_{k}}=f_{X_{k+1}| X_{k}}(x_{k+1} | x_k,\Theta)$, where $f_{X_{k+1}| X_{k}}:\mathbb{R}^d\times\mathbb{R}^d\times B([0,T])\to\mathbb{R}$ and $B([0,T])$ is the space of Borel measurable and bounded (in he supremum norm) functions defined on $[0,T]$ and with values in $\mathbb{R}^s$. Hence, we assume that the transition densities might depend on the parameter function $\Theta:[0,T]\to\mathbb{R}^s$ in the functional way, and this dependence might be nonlinear. By the Markov property, the joint density function is given by
\begin{displaymath}
    f_{(X_1,X_2,\ldots,X_n)}(x_1,x_2,\ldots,x_n,\Theta)=\prod_{k=0}^{n-1}f_{X_{k+1}| X_{k}}(x_{k+1} | x_k, \Theta).
\end{displaymath}
In the pursuit of modeling complex statistical systems, we employ the maximum likelihood based approach (MaxLike) to derive the likelihood function, {see \cite{yoshida1992estimation,bishwal2007parameter}}. The likelihood function, denoted as $\mathcal{L}(\Theta)$, is defined as follows:
    \begin{displaymath}
        \mathcal{L}(\Theta)=-\ln f_{(X_1,X_2,\ldots,X_n)}(x_1,x_2,\ldots,x_n,\Theta)
        =-\sum\limits_{k=0}^{n-1}\ln\Bigl[ f_{X_{k+1}|X_k}(x_{k+1} | x_k,\Theta)\Bigr]
    \end{displaymath}

Here, our aim is to find the optimal values of the model parameters, denoted as $\Theta^*$, that minimize the likelihood function:       
    \begin{displaymath}
    \Theta^*=\hbox{argmin}_{\Theta\in B([0,T])}\mathcal{L}(\Theta).
    \end{displaymath}
In this setting the problem is infinite dimensional and hard to solve numerically. However, such problems might be  tackled with the neural networks. Hence,we reformulate the problem using the maximum likelihood approach to efficiently use the neural network framework.

We can assume (supported by the universal approximation theorem) that the functions can be effectively modeled using neural network architectures. Specifically, we employ a simple feed-forward neural network consisting of affine functions and activation functions. In this context, the parameters of the neural network correspond to its weights. In the Markov setting, as defined in this section, we formulate the neural network maximum likelihood approach (briefly marked as NN-MaxLike) as follows:
    \begin{displaymath}
        \mathcal{L}(w)       =-\sum\limits_{k=0}^{n-1}\ln\Bigl[ f_{X_{k+1}|X_k}(x_{k+1} | x_k,\Theta(w))\Bigr],
    \end{displaymath}
where $\Theta$ is modeled by an artificial neural network equipped with a custom loss function $\mathcal{L}$, and the network's parameters are represented by the weights $w \in \mathbb{R}^N$. It's important to note that we assume that the dimensionality $N$, is strictly smaller than $n$. The specific value of $N$ depends on the architecture of the neural network but is entirely independent of the size of the dataset, $n$.

    Consequently, our objective is to find the optimal set of weights, denoted as $w^*$, that minimizes the customized loss function $w\in\mathbb{R}^N$:

    \begin{displaymath}
w^*=\hbox{argmin}_{w\in\mathbb{R}^{N}}\mathcal{L}(w).
    \end{displaymath}
    Hence, our function that approximates the unknown parameter function $\Theta$ is $\Theta(\ \cdot\ ,w^*)$.
    Now, number of optimization parameters is $O(N)$, which depends on architecture of neural network model but is independent of $n$ (the size of the available data). This innovative NN-MaxLike approach effectively addresses the challenge of high-dimensional parameter optimization, making it a valuable tool in modeling complex systems.

In the framework outlined above, later in the paper we tackle two primary issues. In Section~\ref{section3} we delve into the complexities of multivariate regression, where we explore the modeling of relationships between multiple variables. In Section \ref{section4}, we focus on the challenging task of estimating time-dependent parameters for stochastic differential equations, offering insights and practical guidance for dynamic parameter estimation in evolving systems.

\section{Multivariate regression - heteroscedasticity case}\label{section3}
In this section we firstly derive the formula for the maximum likelihood in the case of multivariate regression with normal residuals. Then we apply the neural network approach to estimate unknown parameters such us covariance matrix. In particular, this generalizes the approach described in Section 4.3.3. in \cite{DSM}, where only the scalar case where discussed.

\subsection{Maximum likelihood for multivariate regression}

In the multivariate regression case we assume $x_1,x_2,\ldots,x_n$ are realizations of the $\mathbb{R}^d$-valued random vectors
\begin{equation}
    X(t_k)=\mu(t_k)+\varepsilon(t_k), \quad k=1,2,\ldots,n,
\end{equation}
at discrete time points $t_k$, $k=1,\ldots,n$,
where $(\varepsilon(t_k))_{k=1,\ldots,n}$ are independent random vectors,  $\varepsilon(t_k)\sim \mathcal{N}(0,\Sigma(t_k))$ with $\Sigma(t_k)\in\mathbb{R}^{d\times d}$, $\mu(t_k)\in\mathbb{R}^d$. We assume that for every $k$ the symmetric covariance matrix $\Sigma(t_k)$ is strictly positive definite.
Here 
$\Theta(t)=[\mu(t),\Sigma(t)]$. Since $(X(t_k))_{k=1,\ldots,n}$ are independent and each $X(t_k)$  has the normal distribution $\mathcal{N}(\mu(t_k),\Sigma(t_k))$, we have that
\begin{eqnarray}
    &&f_{X_{k}|X_{k-1}}(x|x_{k-1},\Theta)=f_{X_k}(x|\Theta)=(2\pi)^{-d/2}(\det(\Sigma(t_k)))^{-1/2}\notag\\
    &&\quad\quad\times\exp\Bigl(-\frac{1}{2}(x-\mu(t_k))^T\Sigma(t_k)^{-1}(x-\mu(t_k))\Bigr),
\end{eqnarray}
for $x\in\mathbb{R}^d$. As it was described in Section \ref{section2} we approximate the parameter function $\Theta=[\mu,\Sigma]$ via the neural network $\Theta(w)=[\mu(w),\Sigma(w)]$. Hence, the loss function has the following form
\begin{equation} 
\label{LOSS_reg}
\mathcal{L}(w)= \frac{1}{2} \sum_{k=1}^{n}  \Bigg[ \ln\Bigl[(2\pi)^d\cdot\det(\Sigma(t_k,w))\Bigr] + (x_k - \mu(t_k,w))^T \cdot\Sigma^{-1}(t_k,w)\cdot (x_k - \mu(t_k,w)) \Bigg],
\end{equation} 
and we are minimizing it in order to obtain the optimal weight for our neural network
\begin{displaymath}
w^*=\hbox{argmin}_{w\in\mathbb{R}^{N}}\mathcal{L}(w).
    \end{displaymath}
In the next section we illustrate this approach with the  numerical example performed for in the two-dimensional setting. 
\subsection{Numerical experiments} 
In Section 4.3.3 of \cite{DSM} the authors provide example with using neural network for one-dimensional nonlinear regression  with nonconstant variance. Using described above generalization, we provide two-dimensional example with two correlated one-dimensional time-series. The process takes two steps -- firstly to generate the synthetic dataset, and then to train the neural network model. The code of the experiments for the regression problem can be found here: \url{https://github.com/mwiacek8/DL-based-estimation/tree/main/regression_2D}.

\noindent \textbf{Step 1. Synthetic data generation.} \\
The problem we resolve is the two dimensional regression. We assume that in each point $t$, the value $X=(X_1, X_2)$ is from a two dimensional normal distribution. 

We generate the data according to the distribution:

\begin{eqnarray}
&& X(t_k)=\mu(t_k)+\varepsilon(t_k),\quad k=0,1,2,\ldots,n,
\end{eqnarray}

\noindent where $\mu(t_k) = [\mu_1(t_k), \mu_2(t_k)]$, $\varepsilon(t_k)$ are independent random variables with the distribution $\mathcal{N}(0,\Sigma(t_k))$, and $t_k = k \frac{2 \pi}{n}$. We simulated $n=3000$ points. We analyze three cases for parameters setting $\Theta(t) = \left[ \mu^1(t), \mu^2(t), \sigma^1(t), \sigma^2(t), \varrho(t)\right]$ and present our results below. 

\begin{figure}[h]
\centering

\includegraphics[width=0.45\linewidth]{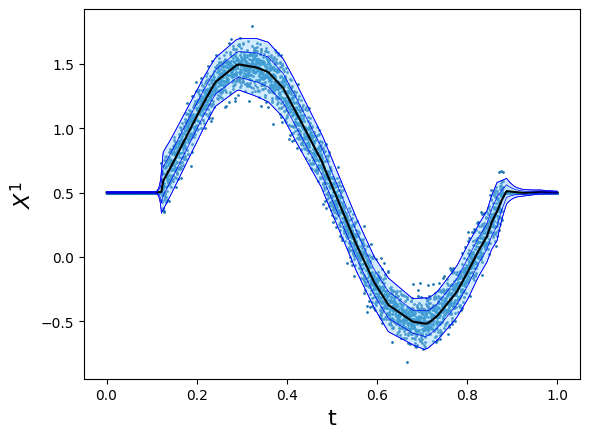}
\includegraphics[width=0.45\linewidth]{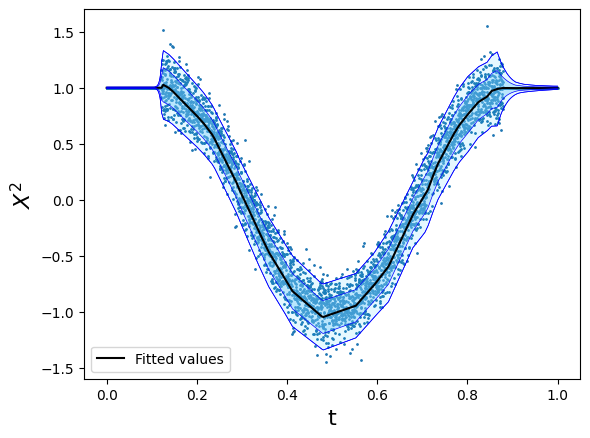}

\caption{Case 1. Fitted values of $\mu_1$ and $\mu_2$ for $X_1$ and $X_2$ respectively, together with 68\% and 95\% prediction intervals}

\label{fig_reg_case1}
\end{figure}

\begin{figure}[h]
\centering
\includegraphics[width=0.45\linewidth]{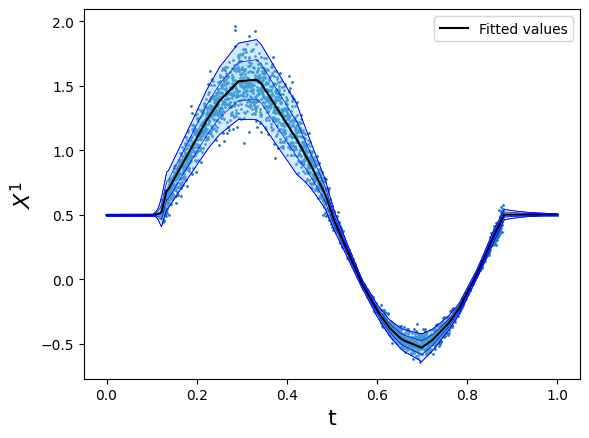}
\includegraphics[width=0.45\linewidth]{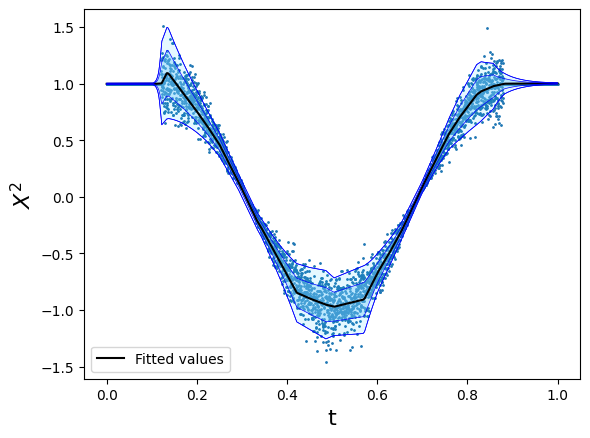}
\caption{Case 2. Fitted values of $\mu_1$ and $\mu_2$ for $X_1$ and $X_2$ respectively, together with 68\% and 95\% prediction intervals}

\label{fig_reg_case2}
\end{figure}

\begin{figure}[h]
\centering
\includegraphics[width=0.45\linewidth]{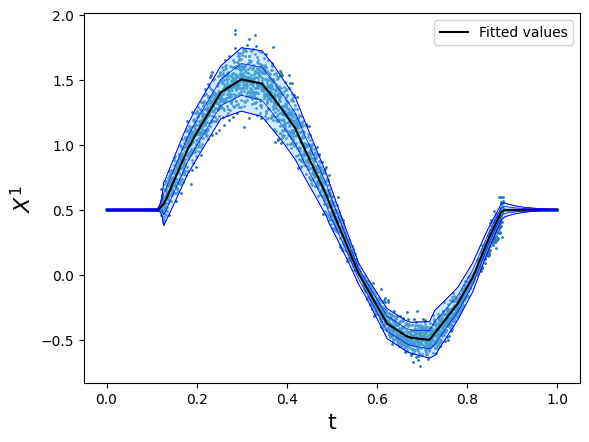}
\includegraphics[width=0.45\linewidth]{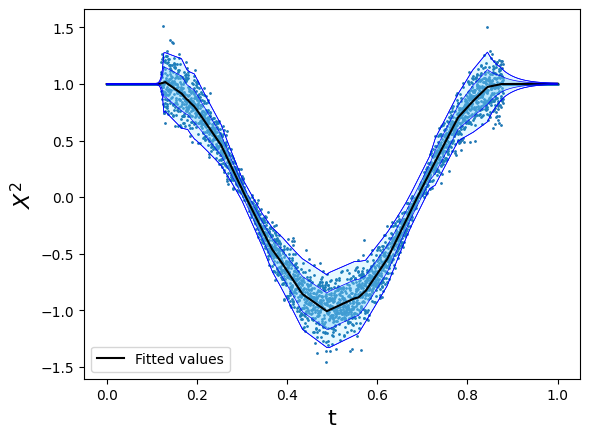}

\caption{Case 3. Fitted values of $\mu_1$ and $\mu_2$ for $X_1$ and $X_2$,respectively, together with 68\% and 95\% prediction intervals}

\label{fig_reg_case3}
\end{figure}

\begin{itemize}
\item \textbf{Case 1 - only mean functions are time-dependent.} \\

\noindent We set the parameters as $\mu_{1}(t) = 0.5 + \sin(t)$, $\mu_{2}(t) = \cos(t)$, $\sigma_1(t)=\sigma_1 \equiv 0.1$, $\sigma_2(t)=\sigma_2 \equiv 0.15$, $\varrho(t) =\varrho
\equiv 0.5$. Our covariance matrix is constant in time and we take  
$$\Sigma_1 = \begin{bmatrix}
\sigma_{1}^{2} & \varrho \sigma_1 \sigma_2 \\
\varrho \sigma_1 \sigma_2  & \sigma_{2}^{2}
\end{bmatrix}.$$ 
Modelled parameter that we want to reproduce with the neural network model is $\mathbb{R}^5\ni\Theta(t) = \left[\mu_1(t), \mu_2(t), \sigma_1, \sigma_2, \varrho \right]$.  \\

\item \textbf{Case 2 - mean functions and variances are time dependent, while correlation is constant.} \\

\noindent In this case we modify our covariance matrix in the following way

$$\Sigma_2(t) = \begin{bmatrix}
\sigma_{1}^{2} |\mu_1(t)|^2 & \varrho \sigma_1 \sigma_2 |\mu_1(t)| \cdot |\mu_2(t)| \\
\varrho \sigma_1 \sigma_2 |\mu_1(t)| \cdot |\mu_2(t)| & \sigma_{2}^{2} |\mu_2(t)|^2
\end{bmatrix},$$
which results in time-dependency for variance parameters, because now the variances $\sigma_1(t) = \sigma_1 |\mu_1(t)|$ and $\sigma_2(t) = \sigma_2 |\mu_2(t)|$ are time-dependent, while again the correlation is constant and equal to  $\varrho(t)=\varrho \equiv 0.5 $. We generate data in the same way as in the previous case and our modelled parameters is $\Theta(t) = \left[\mu_1(t), \mu_2(t), \sigma_1(t), \sigma_2(t), \varrho \right]$ \\

\item \textbf{Case 3 - all parameters are time-dependent. }\\

We assume that the covariance matrix is as follows
$$\Sigma_3(t) = \begin{bmatrix}
\sigma_{1}^{2} |\mu_1(t)| & \varrho \sigma_1 \sigma_2 |\mu_1(t)| \cdot |\mu_2(t)| \\
\varrho \sigma_1 \sigma_2 |\mu_1(t)| \cdot |\mu_2(t)| & \sigma_{2}^{2} |\mu_2(t)|
\end{bmatrix},$$

 then $\sigma_1(t) = \sigma_1 \sqrt{\mu_1(t)}$ and $\sigma_2(t) = \sigma_2 \sqrt{\mu_2(t)}$ and also the correlation is time-dependent because it is now equal to $\varrho(t)=\varrho \sqrt{|\mu_1(t)|} \sqrt{|\mu_2(t)|}  $ with $\varrho=0.5$. So the modelled parameter $$\Theta(t) = \left[ \mu_1(t), \mu_2(t), \sigma_1(t), \sigma_2(t), \varrho(t) \right].$$ 
\end{itemize}
\textbf{Step 2. Creating and training the neural network model.} \\
For $d=2$ the custom loss function  \eqref{LOSS_reg} has the form
\begin{eqnarray*}
    &&\mathcal{L}(w) = \sum_{i=1}^{n}\Biggl[ \ln \biggl( 2\pi \sigma_1(t_i,w) \sigma_1(t_i,w) \sqrt{1-\varrho^2(t_i,w)} \biggr) \notag\\
    &&+ \frac{1}{2(1-\varrho^2(t_i,w))} \biggl( \Bigl( \frac{x_{i}^{(1)} - \mu_1(t_i,w)}{\sigma_1(t_i,w)}\Bigr)^2 + \Bigl(\frac{x_{i}^{(2)} - \mu_2(t_i,w)}{\sigma_2(t_i,w)}\Bigr)^2\notag\\ 
    &&\quad\quad- 2 \varrho(t_i,w) \frac{(x_{i}^{(1)} - \mu_1(t_i,w))(x_{i}^{(2)} - \mu_2(t_i,w))}{\sigma_1(t_i,w)\sigma_2(t_i,w)}\biggr)\Biggr].
\end{eqnarray*}
We construct neural network model with four layers and use our custom loss function for the training. Activation function for the output layer is \textit{linear}, for previous layers we used \textit{ReLu}. Training was conducted on \textit{batch\_size=16} and $1000$ epochs. The results obtained for our data are presented in Figures \ref{fig_reg_case1} - \ref{fig_reg_case3}.

We can observe that presented model is a good fit - not only does it estimate the trend perfectly, but it is also able to properly select the level of uncertainty. Where there is more variation in the data, the confidence intervals are much wider.

\begin{figure}[h]
\centering
\subfloat[][Case 1 - $\varrho$]{
\includegraphics[width=0.25\linewidth]{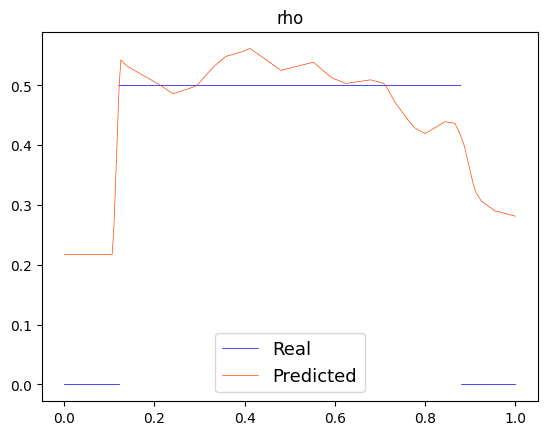}}
\quad
\subfloat[][Case 1 - $\sigma_1$]{
\includegraphics[width=0.25\linewidth]{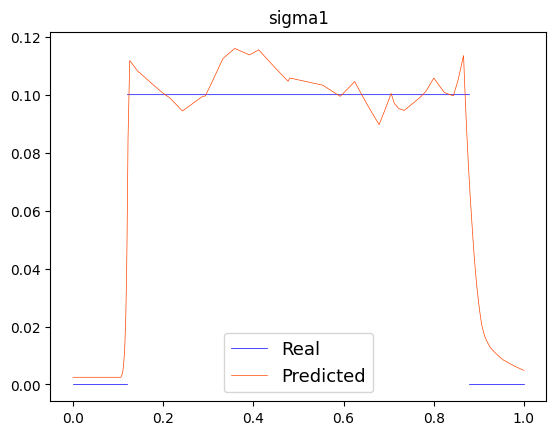}}
\quad
\subfloat[][Case 1 - $\sigma_2$]{
\includegraphics[width=0.25\linewidth]{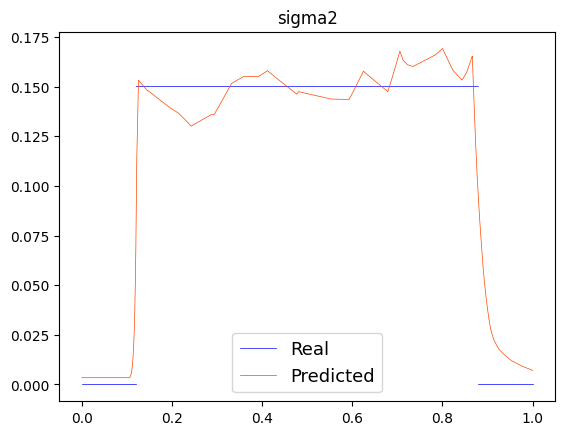}}

\subfloat[][Case 2 - $\varrho$]{
\includegraphics[width=0.25\linewidth]{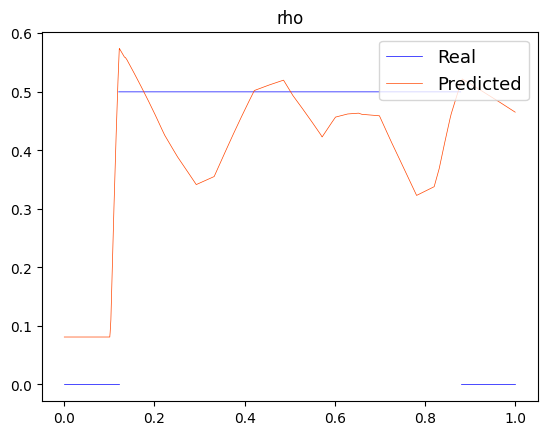}}
\quad
\subfloat[][Case 2 - $\sigma_1$]{
\includegraphics[width=0.25\linewidth]{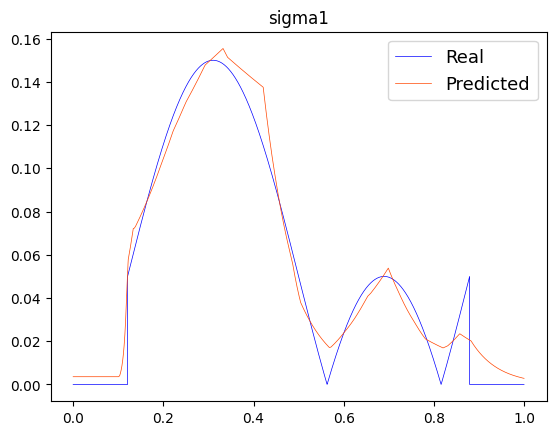}}
\quad
\subfloat[][Case 2 - $\sigma_2$]{
\includegraphics[width=0.25\linewidth]{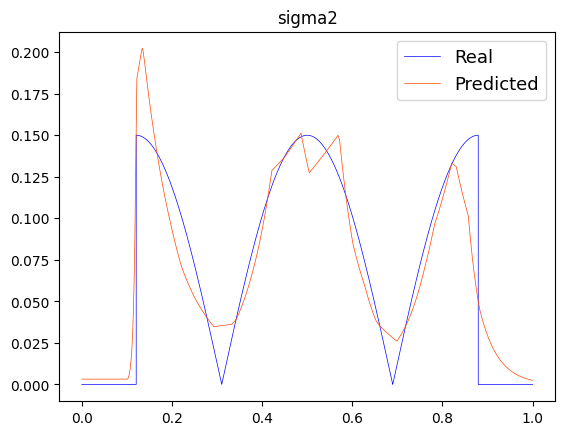}}

\subfloat[][Case 3 - $\varrho$]{
\includegraphics[width=0.25\linewidth]{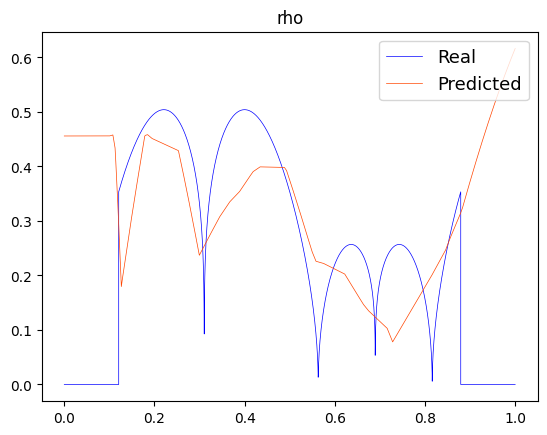}}
\quad
\subfloat[][Case 3 - $\sigma_1$]{
\includegraphics[width=0.25\linewidth]{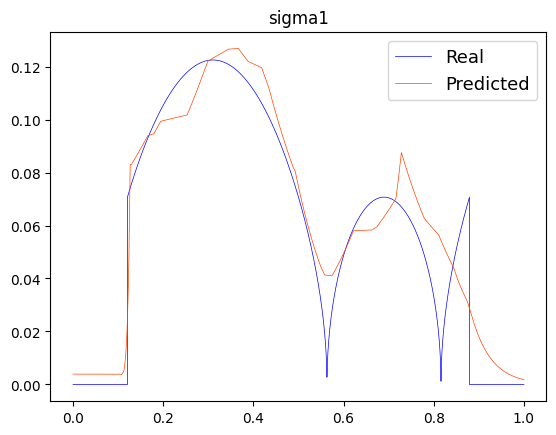}}
\quad
\subfloat[][Case 3 - $\sigma_2$]{
\includegraphics[width=0.25\linewidth]{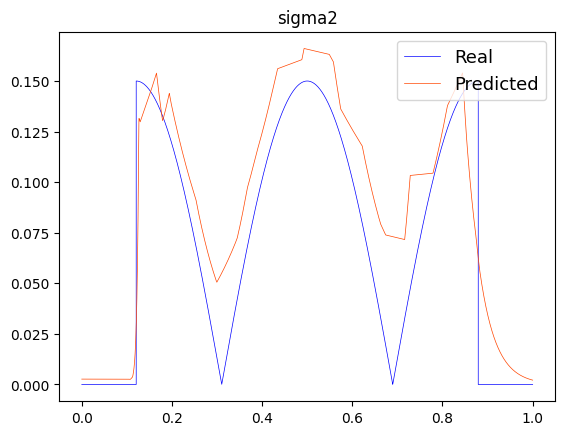}}

\caption{parameters reproduced by the neural network.}

\label{fig_rho_sigmas}
\end{figure}

We can also take a look how our parameters were reproduced by the model (Figure \ref{fig_rho_sigmas}).

\section{Estimation of time-dependent parameters for SDEs-based models}\label{section4}

We now analyze the case, where we assume we have access to a single trajectory of the process $X$. We assume that the process $X$ is a solution of the underlying SDE, and the coefficients $a,b$ involve time-dependent parameters $\Theta=\Theta(t)$. We would like to employ previously derived approach into finding the actual parameter function $\Theta$ of the underlying SDE-based model. 

\subsection{Quasi-Maximum Likelihood approach}
We investigate system of  stochastic differential equations of the following form
\begin{equation}
\label{main_equation_1}
	\left\{ \begin{array}{ll}
	\displaystyle{
	\rd X(t) = a(t,X(t), \Theta(t))\rd t + b(t,X(t), \Theta(t)) \rd W(t), \ t\in [0,T]},\\
	X(0)=x_0, 
	\end{array} \right.
\end{equation}
where parameters function $\Theta: [0,T] \rightarrow \mathbb{R}^s$ is time dependent, and $W$ is an $m$-dimensional Wiener process. We want to estimate $\Theta$ based on discrete realizations $x_0, x_1,\ldots,x_n$ of the process $X(t)$ at $t \in \{t_0, t_1,...,t_n\}$ and $t_i = ih$, $h=T/n$, $i=0,1,...,n$.  Note that the sequence of random variables $(X(t_k))_{k=0,1,\ldots,n}$ has Markov property \eqref{disc_markov_prop}. Therefore the joint density of the vector $(X(t_1),\ldots,X(t_n))$ is given by
\begin{equation}
    f_{(X(t_1),\ldots,X(t_n))}(x_1,\ldots,x_n, \Theta)=\prod_{k=0}^{n-1} f_{X(t_{k+1}) | X(t_k)} (x_{k+1}|x_k, \Theta),
\end{equation}
see, for example, Remark 5.12, page 68 in \cite{SSAS}.

For general $X$ we do not know the transition density function. Moreover, in the case when $\Theta$ depends on $t$ even if the transition density is known such information might be useless, see Remark \ref{nonlin_tr_f_1}. Hence, the exact maximum likelihood is not known. Hence, we use Euler scheme to approximate and its transition density in order to approximate transition density for $X$, see Section 11.4 in \cite{SSAS}.

For the estimation procedure, we impose the following non-degeneracy of diffusion condition:
non-degenerative (ND), the symmetric matrix $(b \cdot b^T)(t, x, z)$ is strictly positive definite for all $(t, x, z) \in [0, T] \times \mathbb{R}^d \times \mathbb{R}^s$.

Recall that the Euler scheme is defined as follows:
\begin{align}
    &X^E(0) = x_0, \notag \\
    &X^E(t_{k+1}) = X^E(t_k) + a(t_k, X^E(t_k), \Theta(t_k))h + b(t_k, X^E(t_k), \Theta(t_k)) \Delta W_k, \\
    &\text{for } k = 0, 1, \ldots, n-1. \notag
\end{align}

The conditional law of $X^E(t_{k+1})$ given $X^E(t_k)$ is the $d$-dimensional normal distribution 
$$X^E(t_{k+1}) | X^E(t_k) \sim \mathcal{N} \Bigg(X^E(t_k) + a(t_k, X^E(t_k), \Theta(t_k))h, h(b\cdot b^T)(t_k, X^E(t_k), \Theta(t_k)) \Bigg).$$
Hence, we have the following transition density for the Euler scheme
\begin{multline*}
f_{X^E(t_{k+1}) | X^E(t_k)} (x|x_k, \Theta) =  (2\pi)^{-d/2}\det\bigl( h (b\cdot b^T) (t_k, x_k, \Theta(t_k)) \Bigr)^{-\frac{1}{2}} \\
\times\exp \Bigl[ -\frac{1}{2h} (x-x_k - a(t_k, x_k, \Theta(t_k))h)^T (b\cdot b^T)^{-1}(t_k, x_k, \Theta(t_k))(x-x_k - a(t_k, x_k, \Theta(t_k))h)\Bigr]
\end{multline*} 
  for $x=[x^1,\ldots,x^d]\in\mathbb{R}^d$, $x_k=[x_k^1,\ldots,x_k^d]\in\mathbb{R}^d$.
Moreover, it is easy to see that  the sequence of random variables $(X^E(t_k))_{k=0,1,\ldots,n}$ has Markov property \eqref{disc_markov_prop}. Hence, \begin{equation}
    f_{(X^E(t_1),\ldots,X^E(t_n))}(x_1,\ldots,x_n,\Theta)=\prod_{k=0}^{n-1} f_{X^E(t_{k+1}) | X^E(t_k)} (x_{k+1}|x_k,\Theta),
\end{equation}
This gives the following negative log-quasi-likelihood  for  $x_k=[x_k^1,\ldots,x_k^d]\in\mathbb{R}^d$
\begin{multline*}
\mathcal{L}(\Theta) = -\sum_{k=0}^{n-1} \ln \Bigg[ f_{X^E(t_{k+1}) | X^E(t_k)} (x_{k+1}|x_k,\Theta) \Bigg] = \\
\frac{1}{2} \sum_{k=0}^{n-1}  \Bigg[ \ln\Bigl[(2\pi)^{d}\det\bigl(  h (b\cdot b^T) (t_k, x_k, \Theta(t_k)) \Bigr)\Bigr] \\
+ \frac{1}{h}(\Delta x_k - a(t_k, x_k, \Theta(t_k))h)^T (b\cdot b^T)^{-1}(t_k, x_k, \Theta(t_k))(\Delta x_k - a(t_k, x_k, \Theta(t_k))h) \Bigg].
\end{multline*} 
As it was described earlier, we approximate $\Theta = \Theta(t)$ by using neural network $\Theta(\ \cdot\ ,w)\approx\Theta(\ \cdot\ )$, where we denote by $w\in\mathbb{R}^N$ the weights of the neural network. The loss function for the neural network is given by

\begin{multline} 
\label{LOSS}
\mathcal{\hat L}(w)= \frac{1}{2} \sum_{k=0}^{n-1}  \Bigg[ \ln\Bigl[(2\pi)^d\det\Bigl( h (b\cdot b^T) (t_k, x_k, \Theta(t_k,w)) \Bigr)\Bigr] \\
+ \frac{1}{h}\Bigl(\Delta x_k - a(t_k, x_k, \Theta(t_k,w))h\Bigr)^T (b\cdot b^T)^{-1}(t_k, x_k, \Theta(t_k,w))\Bigl(\Delta  x_k - a(t_k, x_k, \Theta(t_k,w))h\Bigr) \Bigg],\notag
\end{multline} 
where  $\Delta x_k=x_{k+1}-x_k$. As before, we arrive at optimal weights by minimizing the loss function
\begin{equation}
w^*=\hbox{argmin}_{w\in\mathbb{R}^{N}}\mathcal{\hat L}(w),
\end{equation}
and we take $\Theta(\ \cdot\ ,w^*)$ as the optimal approximation of the real (but unknown) parameter function $\Theta$.
%
%
%
%
%
\begin{remark}
    \label{nonlin_tr_f_1}
    Let us consider the following Ornstein-Uhlenbeck proces with constant $\theta_1$ and varying $\theta_2=\theta_2(t)$
\begin{equation}
\label{OU_1}
    dX(t)=-\theta_1 X(t) \rd t+\theta_2(t)\rd W(t),
\end{equation}
where $\theta_1>0$ and $\theta_2:[0,T]\to [0,+\infty)$, $\Theta=[\theta_1,\theta_2]$. Then the transition density is known and has the following form
\begin{eqnarray}
    &&f_{X(t_{k+1})|X(t_k)}(x|x_k, \Theta)=\Bigl(2\pi\cdot \int\limits_{t_k}^{t_{k+1}}\theta_2^2(u)e^{-2\theta_1\cdot (t_{k+1}-u)}\rd u\Bigr)^{-1/2}\notag\\
    &&\quad\quad\times\exp\Bigl[-\frac{(x-e^{-\theta_1\cdot h}\cdot x_k)^2}{2\int\limits_{t_k}^{t_{k+1}}\theta_2^2(u)\cdot e^{-2\theta_1\cdot (t_{k+1}-u)}\rd u}\Bigr].
\end{eqnarray}
Therefore the joint density of the vector $(X(t_1),\dots,X(t_n))$ is also known. Note however that we cannot use it in order to obtain estimator of $\Theta$, since the dependence of the negative log-likelihood function on $\theta_2$ is of functional and nonlinear nature. Hence, in this case we also need to use quasi-likelihood approach (see \cite{yoshida1992estimation}), despite the fact that the transition density has an explicit closed form.
\end{remark}
\subsection{Mean-square continuous dependence on the parameter function $\Theta$}
\noindent\newline

In this subsection, we establish results on the mean-square continuous dependence of a stochastic differential equation on the parameter function $\Theta$. These results allow us to deduce quantitative estimates for the difference between solutions corresponding to different parameter functions. 

By $\|\cdot\|$ we mean the Euclidean norm in $\mathbb{R}^d$ or $\mathbb{R}^s$, or the Frobenius norm in $\mathbb{R}^{d\times m}$ (the meaning is clear from the context). For $\varrho\ge 0$ we denote by $B(0,\varrho)=\{x\in\mathbb{R}^d \ | \ \|x\|\leq\varrho\}$. For $H\in\mathbb{R}$ we take $H_+=\max\{H,0\}$.

\subsubsection{Case I - continuous drift}
                                                           
\begin{enumerate}
    \item [(A0)] $x_0\in\mathbb{R}^d$,
    \item [(A1)] $a$ is Borel measurable   and for all $(t,z)\in [0,T]\times\mathbb{R}^s$ the function $a(t,\cdot,z):\mathbb{R}^d\to\mathbb{R}^d$ is continuous,
    \item [(A2)] there exists $H\in\mathbb{R}$ such that for all $t\in [0,T],x\in\mathbb{R}^d,y\in\mathbb{R}^d,z\in\mathbb{R}^s$ it holds $$\langle x-y,a(t,x,z) - a(t,y,z)\rangle \leq H (1 + \norm{z}) \|x-y\|^2,$$
    \item [(A3)] for all  $\varrho \in [0,+\infty)$ it holds
    $$ \sup_{(t,x)\in [0,T]\times B(0,\varrho)} \|a(t,x,0)\| < + \infty,$$
    \item [(A4)] there exist $\alpha_1\in (0,1], q\in [1,+\infty),L \in [0,+\infty)$ such that for all $t\in [0,T],x\in\mathbb{R}^d,z_1\in\mathbb{R}^s,z_2\in\mathbb{R}^s$ the following holds $$\|a(t,x,z_1) - a(t,x,z_2)\| \leq L(1+\|x\|^q)\|z_1-z_2\|^{\alpha_1},$$
\end{enumerate}
\begin{enumerate}
    \item [(B1)] $b$ is a Borel measurable function,
        \item [(B2)] there exists $L\in [0,+\infty)$ such that for all $t\in [0,T],x\in\mathbb{R}^d,y\in\mathbb{R}^d,z\in\mathbb{R}^s$ it holds $$\|b(t,x,z) - b(t,y,z)\| \leq L(1+\|z\|)\cdot \|x-y\|,$$
    \item [(B3)] $$\sup\limits_{0\leq t \leq T}\|b(t,0,0)\|<+\infty,$$    
    \item [(B4)] there exist  $\alpha_2\in (0,1],L\in [ 0,+\infty)$ such that for all $t\in [0,T],x\in\mathbb{R}^d,z_1\in\mathbb{R}^s,z_2\in\mathbb{R}^s$ $$\|b(t,x,z_1) - b(t,x,z_2)\| \leq L(1+\|x\|)\cdot\|z_1-z_2\|^{\alpha_2},$$
\end{enumerate}
and finally we assume that
\begin{enumerate}
    \item [(T1)] $\Theta:[0,T]\to\mathbb{R}^s$ is Borel measurable and $$ \sup\limits_{0\leq t\leq T}\|\Theta(t)\|< +\infty. $$
\end{enumerate}
\begin{remark} It follows from Fact \ref{prop_sde_1} that under the assumptions above there exists unique strong solution $X$ to SDE \eqref{main_equation_1}. For example, in the one dimensional case, the functions $a(t,x,z)=-x^3\cdot |z|^{\alpha_1}$, $b(t,x,z)=g(x)\cdot |z|^{\alpha_2}$, where $g:\mathbb{R}\to\mathbb{R}$ is any globally Lipschitz mapping, $\alpha_1,\alpha_2\in (0,1]$, satisfy the assumptions (A1)--(A4), (B1)--(B4).
\end{remark}
The main result of this section is as follows.
\begin{theorem}
\label{dist_X1X2}
    Let us assume that $\Theta_i:[0,T]\to\mathbb{R}^s$, $i=1,2$, satisfy the assumption $(T1)$, $a:[0,T]\times\mathbb{R}^d\times\mathbb{R}^s\to\mathbb{R}^d$ satisfies the assumptions (A1)-(A4), and $b:[0,T]\times\mathbb{R}^d\times\mathbb{R}^s\to\mathbb{R}^{d\times m}$ satisfies the assumptions (B1)--(B4). Then
    \begin{equation}
        \sup\limits_{0\leq t\leq T}\Bigl(\mathbb{E}\|X_1(t)-X_2(t)\|^2\Bigr)^{1/2}\leq C\Bigl(\|\Theta_1-\Theta_2\|^{\alpha_1}_{\infty}+\|\Theta_1-\Theta_2\|^{\alpha_2}_{\infty}\Bigr),
    \end{equation}
where $X_i=X(x_0,a,b,\Theta_i)$, $i=1,2$, is the unique strong solution of 
\begin{equation}
\label{main_equation_11}
	\left\{ \begin{array}{ll}
	\displaystyle{
	\rd X_i(t) = a(t,X_i(t), \Theta_i(t))\rd t + b(t,X_i(t), \Theta_i(t)) \rd W(t), \ t\in [0,T]},\\
	X_i(0)=x_0,
	\end{array} \right.
\end{equation}
and
\begin{eqnarray}
    C=\max\{C_1,C_2\},
\end{eqnarray}
\begin{eqnarray}
    C_1=L\Bigl(\mathbb{E}(1+\|X_2\|^q_{\infty})^2\Bigl)^{1/2}\cdot\frac{\exp\Bigl(TH_+(1+\|\Theta_2\|_{\infty})+TL^2(1+\|\Theta_1\|_{\infty})^2\Bigr)-1}{ H_+(1+\|\Theta_2\|_{\infty})+L^2(1+\|\Theta_1\|_{\infty})^2},
\end{eqnarray}
\begin{eqnarray}
    C_2=2T^{1/2}L(1+\mathbb{E}[\|X_2\|^2_{\infty}])^{1/2}\cdot\exp\Bigl(TH_+(1+\|\Theta_2\|_{\infty})+TL^2(1+\|\Theta_1\|_{\infty})^2\Bigr).
\end{eqnarray}
\end{theorem}
For the proof we refer to Appendix B.
\begin{remark}
We stress that $C$ in Theorem \ref{dist_X1X2} does not depend on $X_1$. Therefore, if $\|\Theta_1~-~\Theta_2\|_{\infty}~\to~0$ (i.e. $\Theta_1\to\Theta_2$ uniformly on $[0,T]$) then 
\begin{equation}
    \sup\limits_{0\leq t\leq T}\Bigl(\mathbb{E}\|X(x_0,a,b,\Theta_1)(t)-X(x_0,a,b,\Theta_2)(t)\|^2\Bigr)^{1/2}\to 0,
\end{equation}
and, in particular, $X(x_0,a,b,\Theta_1)(T)\to X(x_0,a,b,\Theta_2)(T)$ in law.
\end{remark}

We will approximate function $\Theta_1$ using neural network $\Theta(\cdot,w)$, where $w$ is correspondingly large vector of weights (depends on particular network architecture), i.e.
$\Theta_1(\cdot) \approx \Theta(\cdot,w).$
\subsubsection{Case II - discontinuous drift}

In the following scalar case ($m=d=1$) with additive noise 
\begin{equation}
\label{main_equation_2}
	\left\{ \begin{array}{ll}
	\displaystyle{
	\rd X(t) = a(t,X(t)) \rd t + \Theta(t) \rd W(t), \ t\in [0,T]},\\
	X(0)=x_0, 
	\end{array} \right.
\end{equation}
we consider the following {\it Zvonkin-type} conditions  for $a:[0,T]\times\mathbb{R}\to\mathbb{R}$ and $\Theta:[0,T]\to\mathbb{R}$:
\begin{enumerate}
    \item [(C0)] $x_0\in\mathbb{R}$,
    \item [(C1)] $a$ is Borel measurable,
    \item [(C2)] there exists $H\in\mathbb{R}$ such that for all $t\in [0,T]$, $x,y\in\mathbb{R}$ it holds
    \begin{equation}
        (x-y)(a(t,x)-a(t,y))\leq H|x-y|^2,
    \end{equation}
    \item [(C3)] there exists $D\in [0,+\infty)$ such that for all $t\in [0,T],x\in\mathbb{R}$ it holds
    \begin{equation}
        |a(t,x)|\leq D,
    \end{equation}
    \item [(D1)] $\Theta$ is Borel measurable,
    \item [(D2)] there exist $d_0,d_1\in (0,+\infty)$ such that for all $t\in [0,T]$
    \begin{equation}
        d_0\leq \Theta(t) \leq d_1. 
    \end{equation}
\end{enumerate}
It follows from Theorem 4 in \cite{Zvon74} that under the assumptions (C0)--(C2), (D1), (D2) the SDE \eqref{main_equation_2} has a unique strong solution such that
\begin{equation}
\label{X_Sol_zv_est}
    \mathbb{E}\Bigl(\sup\limits_{0\leq t\leq T}|X(t)|^2\Bigr)<+\infty,
\end{equation}
see also page 396 in \cite{KarShr}. For example, the function $a(t,x)=\mu-\kappa\cdot\sign(x)$, with $\mu\in\mathbb{R}$ and $\kappa\geq 0$, satisfies the assumptions (C1)--(C3).
\begin{theorem}
\label{dist_X1X2_discont}
    Let us assume that $\Theta_i:[0,T]\to\mathbb{R}$, $i=1,2$, satisfy the assumptions $(D1),(D2)$, and $a:[0,T]\times\mathbb{R}\to\mathbb{R}$ satisfies the assumptions (C1)-(C3). Then
    \begin{equation}
        \sup\limits_{0\leq t\leq T}\Bigl(\mathbb{E}|X_1(t)-X_2(t)|^2\Bigr)^{1/2}\leq e^{TH_+ }\|\Theta_1-\Theta_2\|_{L^2[0,T]},
    \end{equation}
where $X_i=X(x_0,a,\Theta_i)$, $i=1,2$, is the unique strong solution of 
\begin{equation}
\label{main_equation_12}
	\left\{ \begin{array}{ll}
	\displaystyle{
	\rd X_i(t) = a(t,X_i(t))\rd t + \Theta_i(t) \rd W(t), \ t\in [0,T]},\\
	X_i(0)=x_0.
	\end{array} \right.
\end{equation}
\end{theorem}
For the proof see Appendix \ref{appendixB}.
%
\subsection{Numerical experiments}
\noindent\newline
In this section, we highlight that existing networks demonstrate proficient estimation of $\sigma$ associated with diffusion. However, challenges arise in parameter estimation related to drift, despite the loss function encompassing parameters from both drift and diffusion. Addressing the estimation of drift parameters concurrently with the estimation of time-dependent parameters in diffusion is identified as a potential area for future work. The code of the experiments for the SDEs problems can be found here: \url{https://github.com/mwiacek8/DL-based-estimation/tree/main/SDEs}.

We present results for estimating unknown time-dependent parameters in SDEs-based models. There are three examples presented. The proposed approach enables us to approximate the underlying SDE based on the single trajectory by deep learning. We only need to assume the SDE type, which means specific coefficients, and generate one single trajectory (which serves, in numerical experiments, as synthetically generated data). An approach with access to a single trajectory is consistent with real-world problems, where we typically do not have access to the SDEs coefficients but we can observe single time series data. 

% ...




\subsubsection{Problems}
For the test purposes, we analyze the following calibration problems of form \eqref{main_equation_1}:



\begin{itemize}

\item \textbf{Example 1 - Ornstein-Uhlenbeck process}
 \begin{equation}
\label{example1}
d X(t) = \kappa(\mu-X(t)) d t + \sigma(t) d W(t)
\end{equation}

\noindent with $\kappa = 2$, $\mu = 0.5$ and time-dependent $\sigma(t) = 2t + 0.4 + 1.5 \cdot \sin(4t).$ We estimate $\Theta(t) = \sigma(t)$. The coefficients considered satisfy assumptions of Theorem \ref{dist_X1X2}.\\

\item \textbf{Example 2 - discontinuous Ornstein-Uhlenbeck process}

We consider the following SDE (also known as threshold diffusion)
\begin{equation}
\label{example2}
d X(t) = \Big( \mu - \kappa \cdot \sign(X(t)) \Big) d t + \sigma(t) d W(t),
\end{equation}

\noindent with $\kappa = 2$, $\mu = 0.5$ and $\sigma(t) = 2t + 0.4 + 1.5 \cdot \sin(4t)$. By $\sign$ we mean the signum function. Estimated parameter is $\Theta(t) = \sigma(t)$. As previously, we assume that $\kappa$, $\mu$ are known. Note that for the coefficients of the equation \eqref{example2} the assumptions of Theorem~\ref{dist_X1X2_discont} are satisfied.\\

\item \textbf{Example 3 - SDE with nonlinear coefficients}

\begin{equation}
\label{example3}
d X(t) =  \kappa \cdot \cos(X(t))  d t + \Big( \big( \sin(X(t)) + 1.5 \big) \cdot \sigma(t) + 2 \Big) d W(t),
\end{equation}

\noindent with $\kappa = 0.4$ and $\sigma(t) = 2 \cdot \sin(2 \pi t) +t.$ Our parameter  is $\Theta(t) = \sigma(t)$. Again, we assume that $\kappa$ is known to the algorithm. Moreover, the coefficients considered satisfy assumptions of Theorem \ref{dist_X1X2}. \\

\item \textbf{Example 4 - transformed Black-Scholes}
\begin{equation}
\label{example4}
d Y(t) = \mu(t) Y(t) d t + \sigma(t) Y(t) d W(t)
\end{equation}
and its logarithmized version $X(t)=\ln (Y(t))$
\begin{equation}
\label{example4_log}
d X(t) = \Big( \mu(t) - \frac{1}{2} \sigma(t)^2 \Big) d t + \sigma(t) d W(t),
\end{equation}

\noindent with $\mu(t) = 7.5 \cdot t^{2} \sin(3.5\cdot t \cdot \pi)$ and $\sigma(t) = \frac{1}{40} (3 + 3 \cdot t^{2} - 3 \cdot t \cdot \sin(3 \cdot t \cdot \pi)) $. The vector of unknown parameters is taken as $\Theta(t) = [\mu(t), \sigma(t)]$.  

\end{itemize}

For each analyzed example, we generate a single trajectory, and based on that, we approximate time-dependent coefficients through the maximum likelihood approach with deep learning used for solving the optimization problem. We implement suitable loss function with regard to $\Theta$ and train neural network using TensorFlow framework to optimize this loss function. 

\subsubsection{Estimation of chosen parameters}
\begin{figure}[h]
\centering
\subfloat[][Example 1]{
\includegraphics[width=0.20\linewidth]{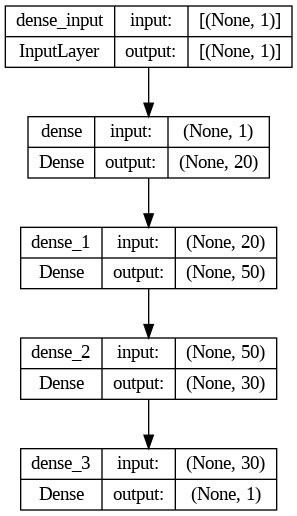}}
\quad
\subfloat[][Example 2]{
\includegraphics[width=0.20\linewidth]{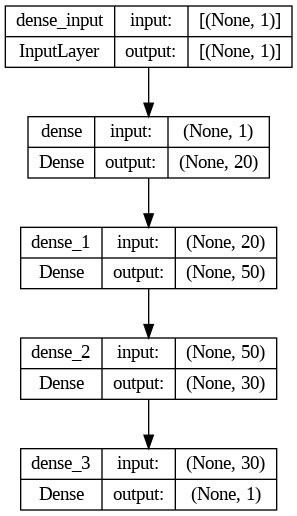}}
\quad
\subfloat[][Example 3]{
\includegraphics[width=0.20\linewidth]{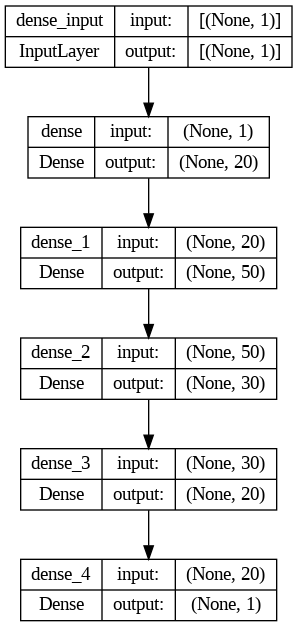}}
\quad
\subfloat[][Example 3]{
\includegraphics[width=0.20\linewidth]{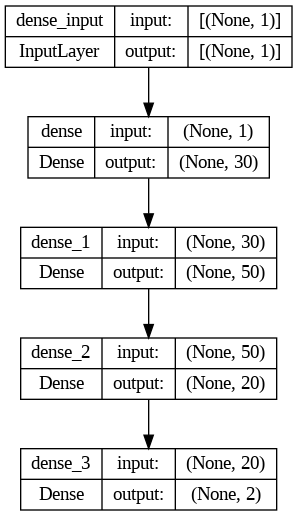}}

\caption{Plot of neural network model graph for each example.}.

\label{fig_nn_mod}
\end{figure}

 We create an appropriate fully-connected neural network model to obtain the approximate values of $\Theta$ (detailed architecture presented on Figure \ref{fig_nn_mod}). The architecture was chosen based on multiple experiments, the objective was high accuracy and stability of obtained results (observed as stability of training procedure). 

\begin{figure}[h]
\centering
\subfloat[][\footnotesize{Example 1. $T = 2.0$. Training parameters: batch\_size=$64$, epochs=$1000$.}]{
\includegraphics[width=0.45\linewidth]{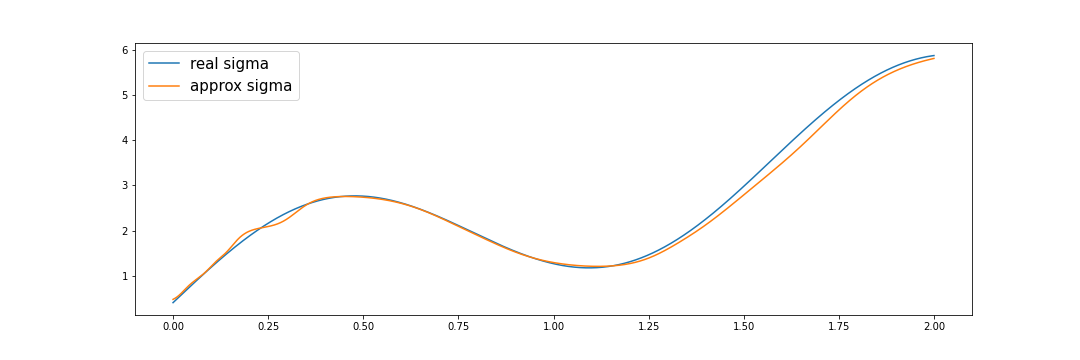}}
\quad
\subfloat[][\footnotesize{Example 2. $T = 2.0$. Training parameters: batch\_size=$64$, epochs=$100$.}]{
\includegraphics[width=0.45\linewidth]{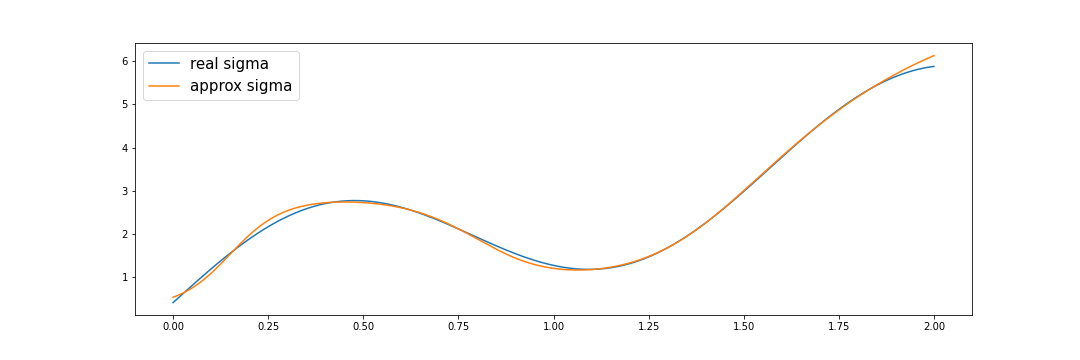}}
\quad
\subfloat[][\footnotesize{Example 3. $T = 3.8$. Training parameters: batch\_size=$4$, epochs=$1050$.}]{
\includegraphics[width=0.45\linewidth]{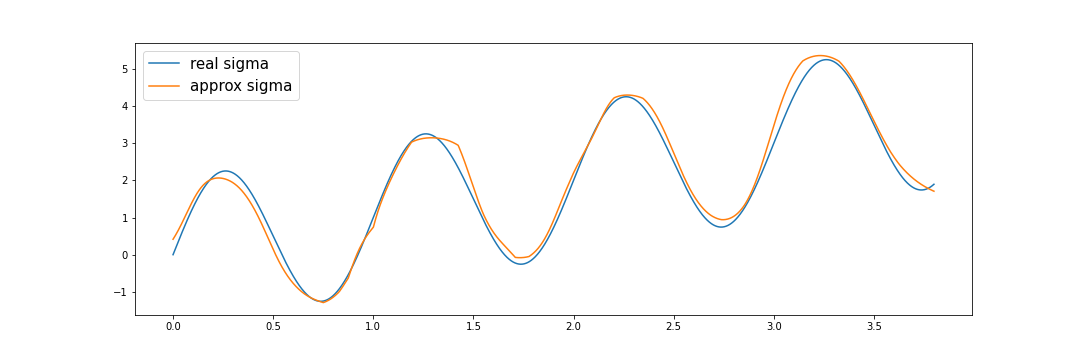}}
\quad
\subfloat[][\footnotesize{Example 4. $T = 1.2$. Training parameters: batch\_size=$228$, epochs=$1000$.}]{
\includegraphics[width=0.45\linewidth]{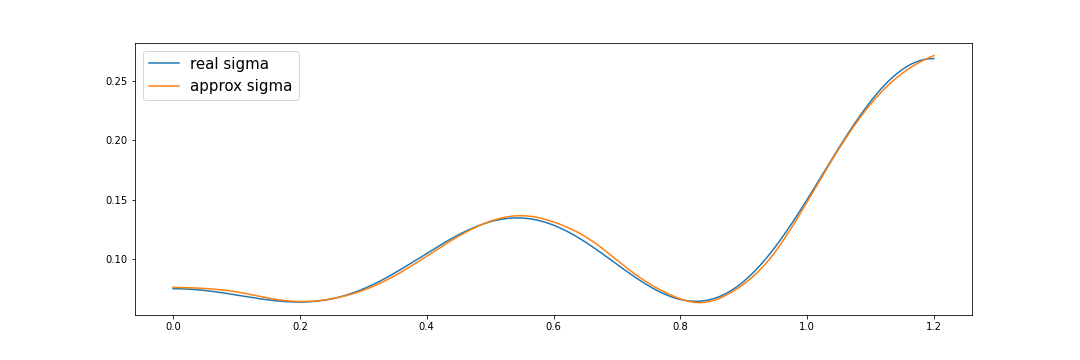}}

\caption{Parameter $\sigma(t)$ for $t \in [0,T]$ estimated with neural network for three given problems.}

\label{fig_sigma}
\end{figure}

\begin{figure}[h]
\centering
\subfloat[][\footnotesize{Example 1. $X_0 = 1.0$, $T=2.0$, $h=0.0002$.} ]{
\includegraphics[width=0.45\linewidth]{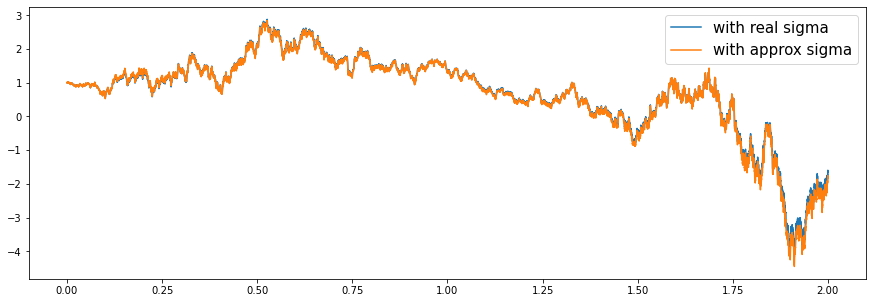}}
\quad
\subfloat[][\footnotesize{Example 2. $X_0 = 1.0$, $T=2.0$, $h=0.0002$.}]{
\includegraphics[width=0.45\linewidth]{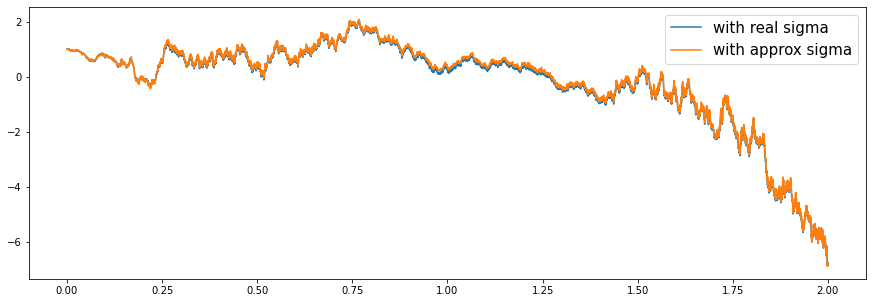}}
\quad
\subfloat[][\footnotesize{Example 3. $X_0 = 1.2$, $T=3.8$, $h=0.00038$.}]{
\includegraphics[width=0.45\linewidth]{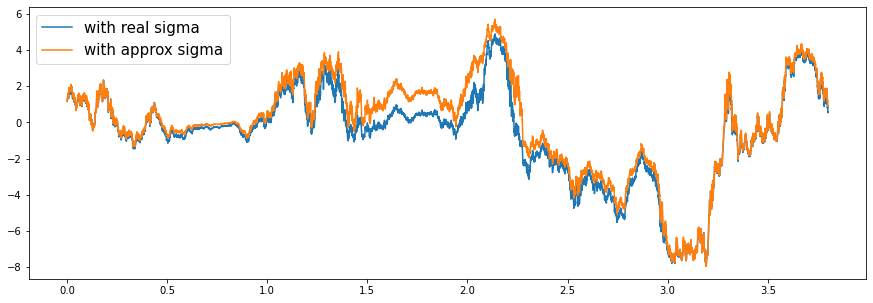}}
\quad
\subfloat[][\footnotesize{Example 4. $X_0 = 1.2$, $T=1.2$, $h=0.000015$.}]{
\includegraphics[width=0.45\linewidth]{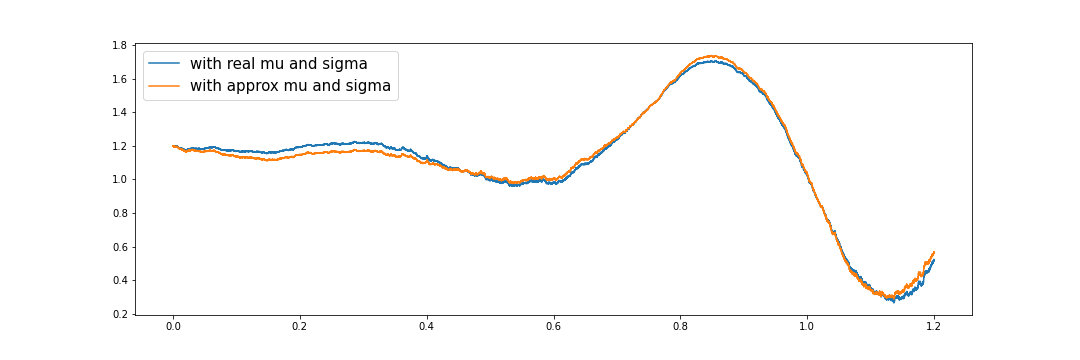}}

\caption{Trajectories $X(t)$ for $t \in [0,T]$ with real and approximated parameters - generated by the Euler-Maruyama scheme with the step-size $h$.}

\label{fig_traj}
\end{figure}

Figure \ref{fig_sigma} presents a comparison between the approximated values of $\sigma(t)$ (denoted as $\hat{\sigma}(t)$) and its real values. Additionally, Figure \ref{fig_traj} displays sample trajectories generated using both exact and approximated parameters, denoted respectively as $X = X(t, \Theta)$ and $\hat{X} = X(t, \hat{\Theta})$ The plots confirm that our approach successfully approximates the real values, validating the reliability of our methodology.

\begin{figure}[h]
\centering
\subfloat[][\footnotesize{Example 1.} ]{
\includegraphics[width=0.45\linewidth]{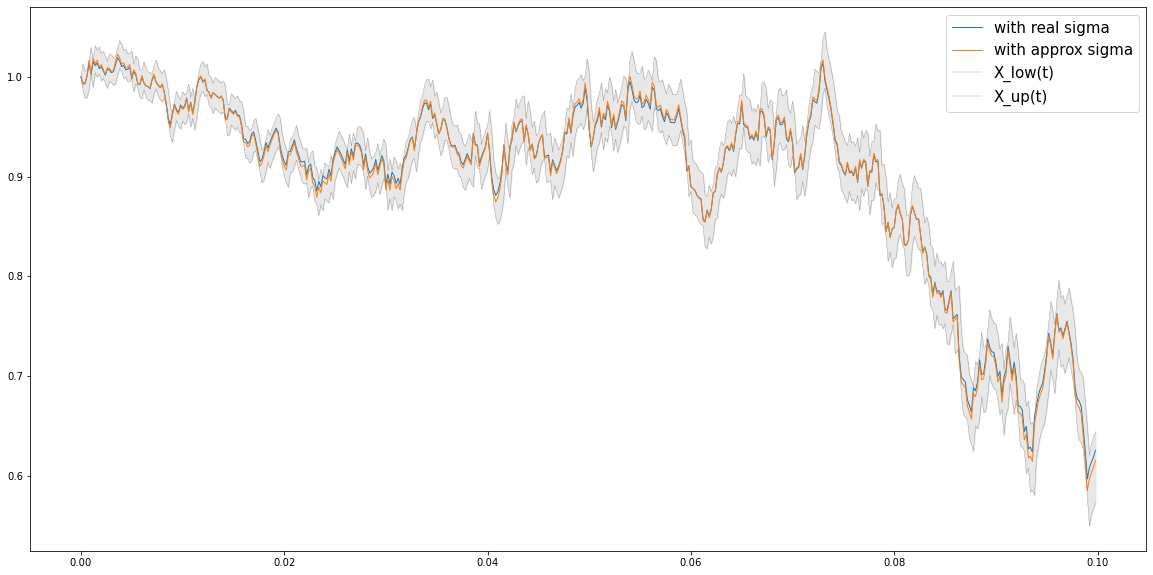}}
\quad
\subfloat[][\footnotesize{Example 2.}]{
\includegraphics[width=0.45\linewidth]{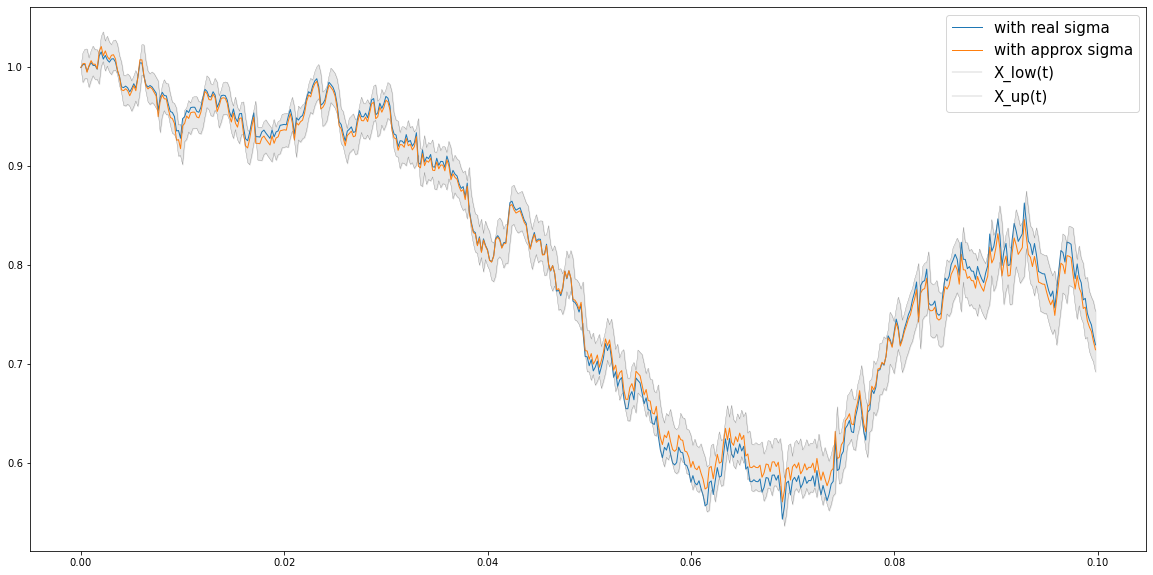}}
\quad
\subfloat[][\footnotesize{Example 3.}]{
\includegraphics[width=0.45\linewidth]{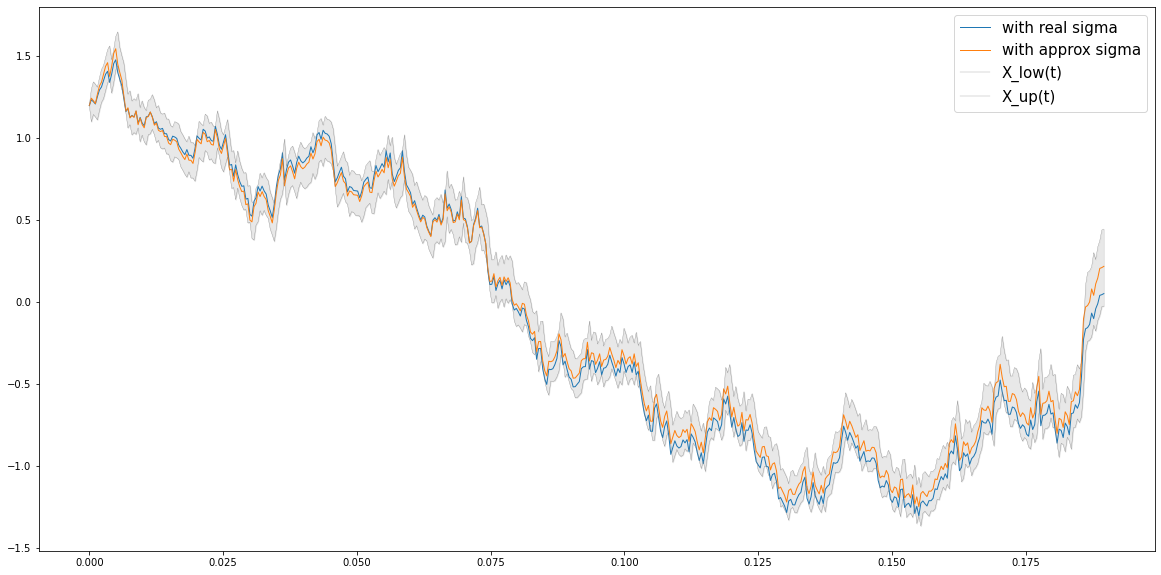}}
\quad
\subfloat[][\footnotesize{Example 4.}]{
\includegraphics[width=0.45\linewidth]{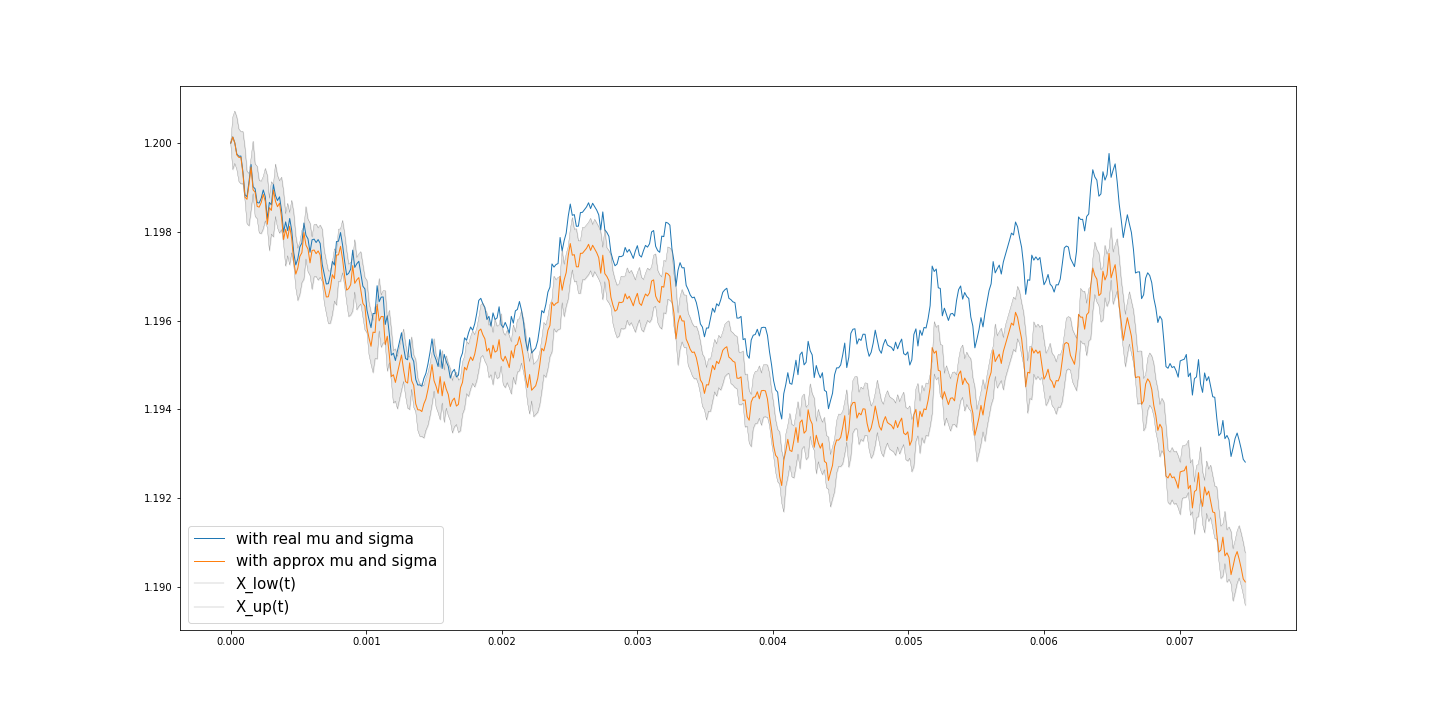}}

\caption{Trajectories $X(t)$ for first $500$ observations with real and approximated parameters with prediction intervals.}

\label{intervals_first}
\end{figure}

\begin{figure}[h]
\centering
\subfloat[][\footnotesize{Example 1.} ]{
\includegraphics[width=0.45\linewidth]{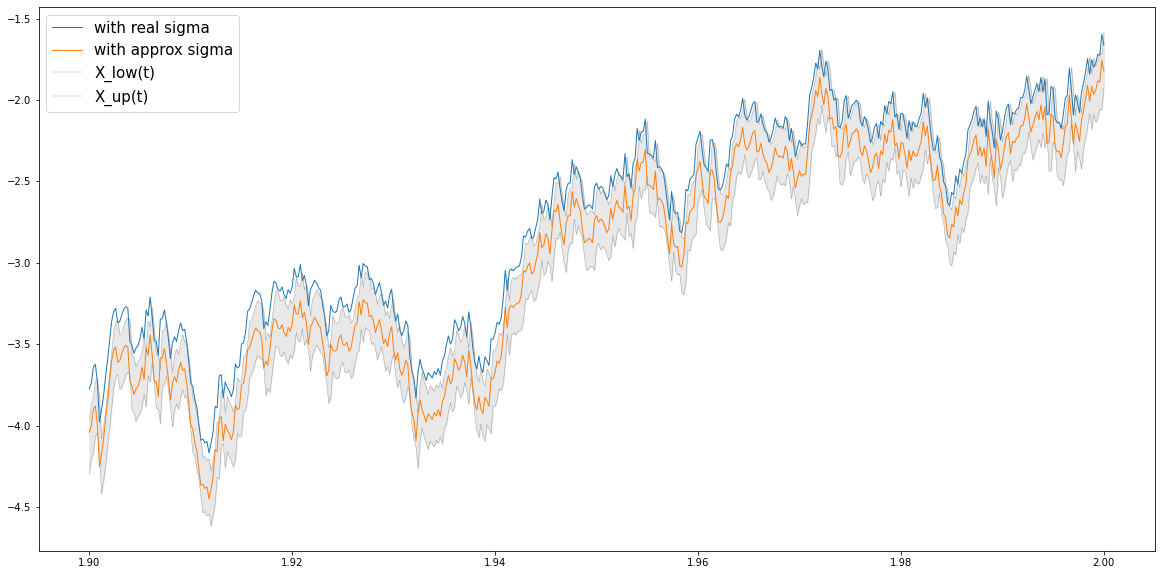}}
\quad
\subfloat[][\footnotesize{Example 2.}]{
\includegraphics[width=0.45\linewidth]{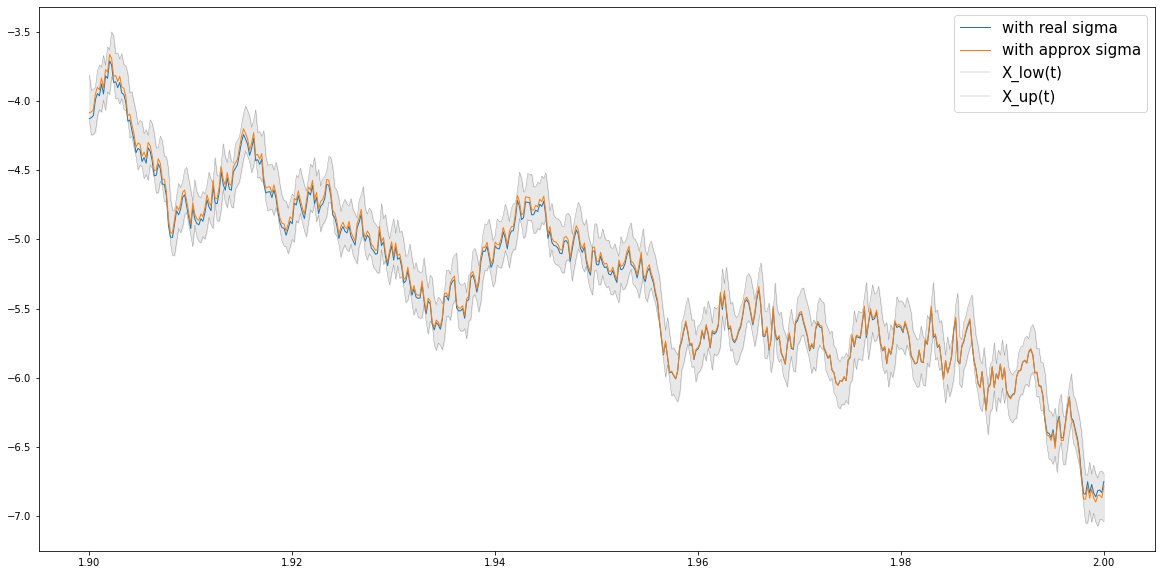}}
\quad
\subfloat[][\footnotesize{Example 3.}]{
\includegraphics[width=0.45\linewidth]{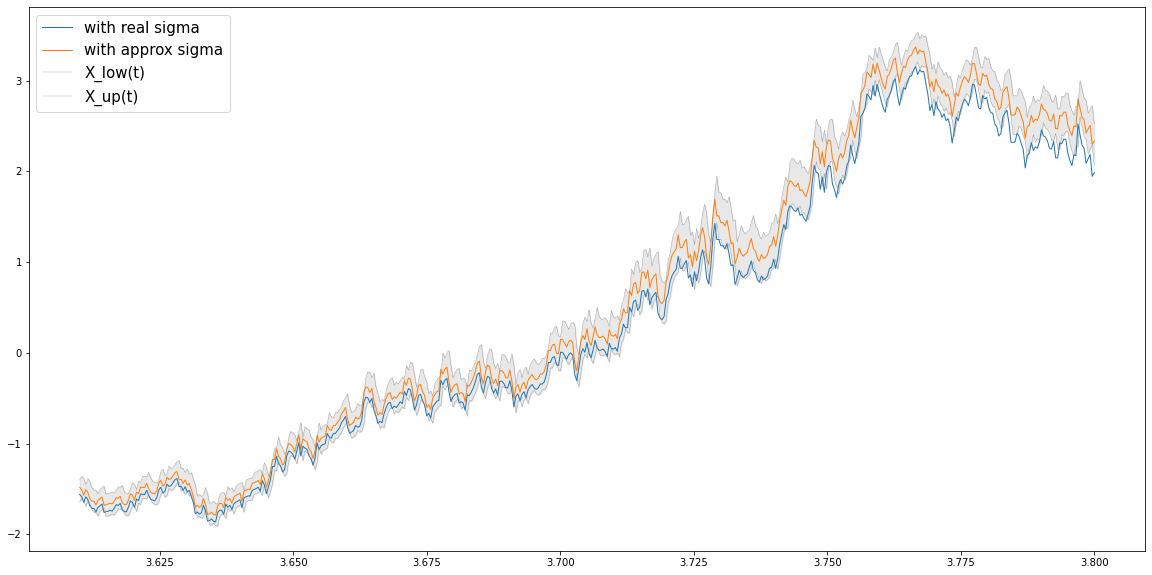}}
\quad
\subfloat[][\footnotesize{Example 4.}]{
\includegraphics[width=0.45\linewidth]{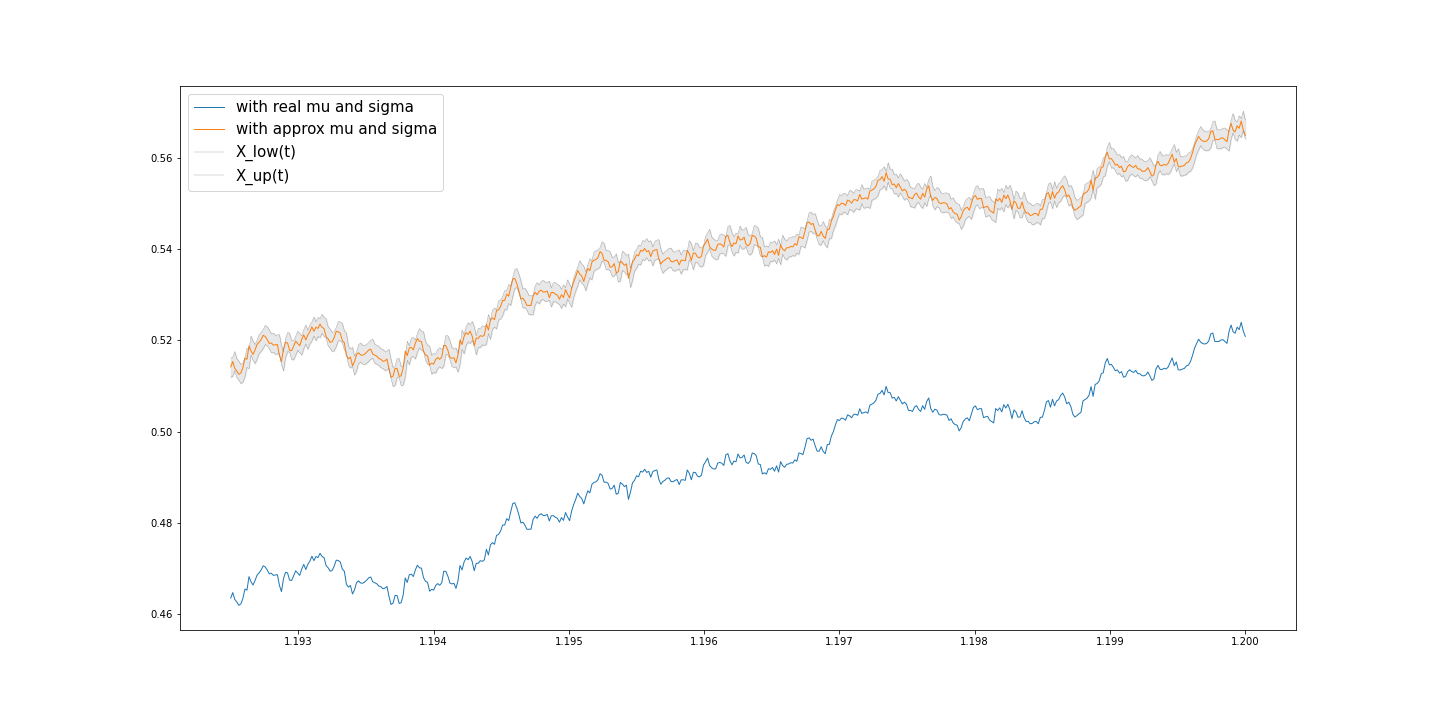}}

\caption{Trajectories $X(t)$ for last $500$ observations with real and approximated parameters with prediction intervals.}

\label{intervals_last}
\end{figure}

To further evaluate the accuracy of our predictions, we analyze $95\%$ prediction intervals, computed as it is described in the Appendix C together with the procedure of Monte Carlo forecasting for SDEs. Figure \ref{intervals_first} illustrates prediction intervals for the initial segment of the trajectories (first $500$ observations for clarity). We observe that in Examples $1$-$3$, the real values closely align with the prediction intervals. In Example $4$ (log-transformed Black-Scholes model), some real values fall outside the intervals, primarily due to the presence of very small $\hat{\sigma}(t)$ values, resulting in narrow intervals.
Continuing the analysis, Figure \ref{intervals_last} depicts prediction intervals for the last $500$ values of the trajectories. We observe a similar trend as before, with Example $4$ showing deviations from the intervals. Despite this, our methodology still yields valuable and insightful approximations for all examples.

To comprehensively evaluate the overall prediction fit, we employ the Mean Squared Error ($MSE$) and $R^2$ metrics. The results, presented in Table \ref{metrics}, indicate that for Examples $1$, $2$, and $4$, the $MSE$ is below $0.5\%$, and the $R^2$ value is above $99.3\%$, signifying an exceptional level of accuracy. In Example $3$, although the values are slightly lower, the $R^2$ score remains notably high at $92.5\%$.



\begin{figure}[h]
\centering
\subfloat[][Example 2 - estimation of $\mu(t)$]{
\includegraphics[width=0.45\linewidth]{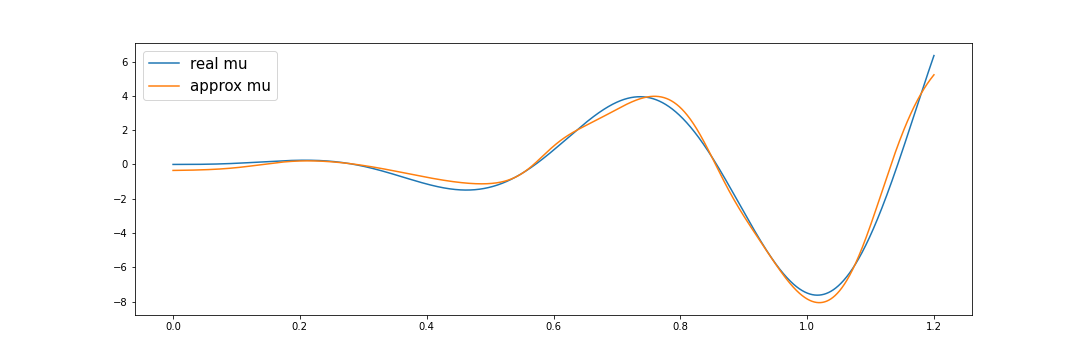}}
\caption{Parameter $\mu(t)$ estimated with our methods for Example 4.}.

\label{fig_bs_m}
\end{figure}


\begin{table}
\begin{tabular}{ |c|c|c|c|c| }
 \hline
& Example 1 & Example 2 & Example 3 & Example 4 \\
 \hline
MSE  & 0.0042 & 0.0035 & 0.4954 & 0.0007 \\
 \hline
$R^2$ & 0.9972 & 0.9987 & 0.9246 & 0.9934 \\
 \hline
\end{tabular} \caption{\label{metrics} MSE and $R^2$ for real and predicted values.} 
\end{table}

\subsubsection{Evaluation of the results}
\label{Evaluation_of_the_results}
To evaluate our methodology, we need to compare the trajectories $X$ and $\hat{X}$ generated with two sets of coefficients - real and approximated by the neural network. Proposed approach is to simulate multiple trajectories for each option and compare distribution of value $X(T)$ with distribution of $\hat{X}(T)$. Our aim is to verify that the distributions are similar. Firstly, we present descriptive statistics to visually inspect how our sample data look like. To obtain quantitative comparison, we use Kolmogorov-Smirnov test.

\subsubsection{\textbf{Descriptive statistics}} \label{desc_stat}
\begin{figure}[h]
\centering
\subfloat[][Example 1]{
\includegraphics[width=0.45\linewidth]{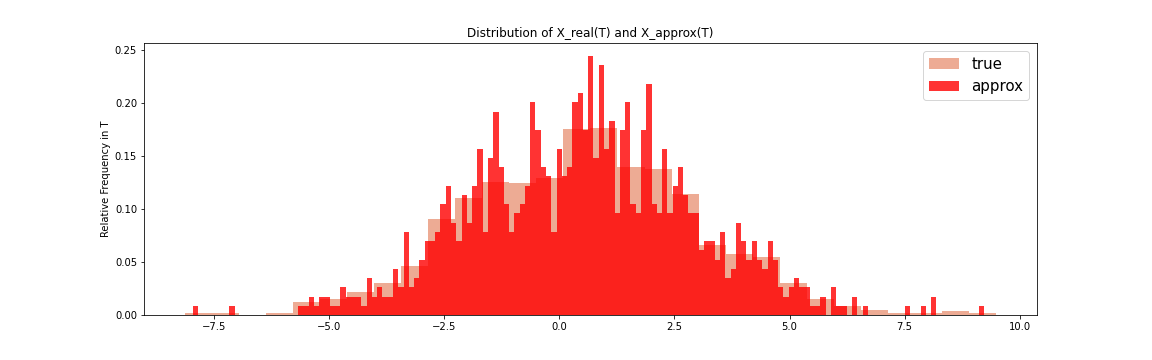}}
\subfloat[][Example 2]{
\includegraphics[width=0.45\linewidth]{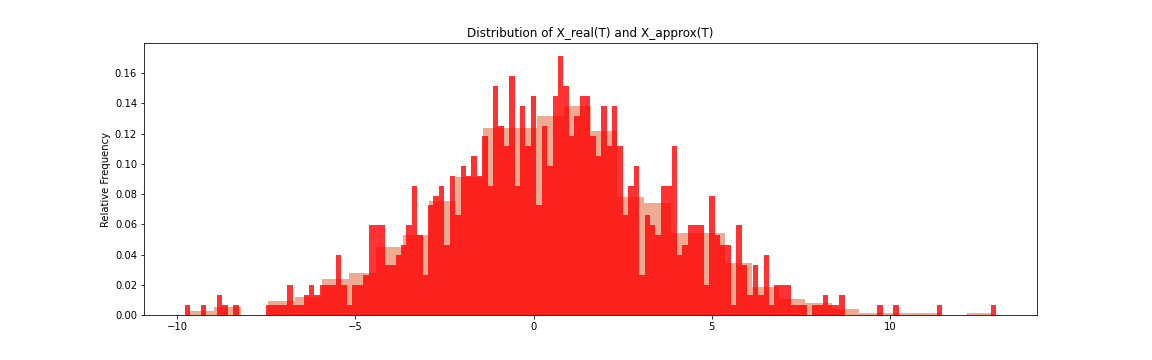}}
\quad
\subfloat[][Example 3]{
\includegraphics[width=0.45\linewidth]{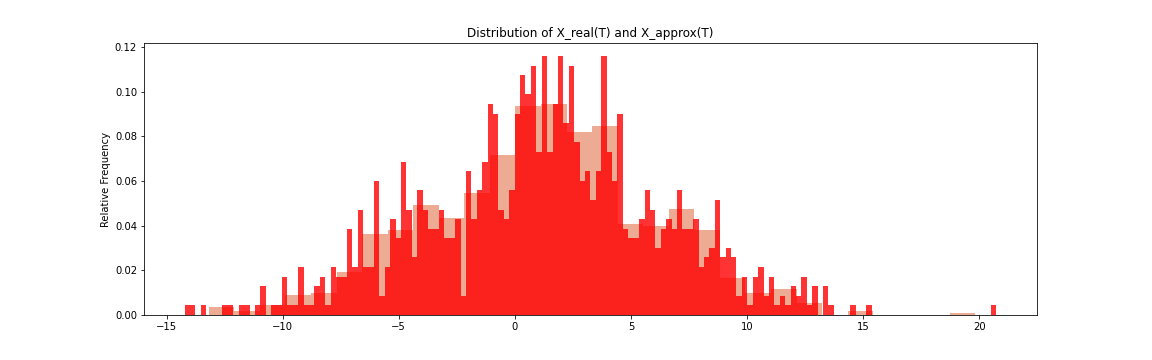}}
\quad
\subfloat[][Example 4]{
\includegraphics[width=0.45\linewidth]{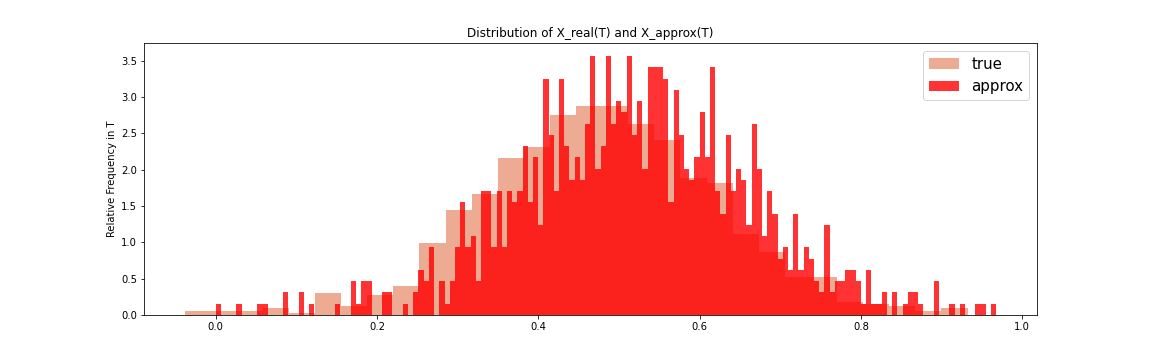}}

\caption{Histogram of values for real $X(T)$ and its approximation with computed $\sigma(t)$. $N=1000$.}

\label{fig_histograms}
\end{figure}

In Figure \ref{fig_histograms} we present histograms generated for values from $N=1000$ number of trajectories $X_{1}, X_{2}, ..., X_{N}$ (marked as "true") and $\hat X_{1}, \hat X_{2}, ..., \hat X_{N}$ (marked as "approx"). It can be seen that the observed values are similar in each case. In Table \ref{histogram_values} we present empirical distributions of the parameters. It is visible that the distributions of trajectories generated using real and neural network approximated coefficients return similar outputs. In the case of a mean value, the biggest difference is for Example $3$ and it is $0.013$. For Example $1$ we notice the same value (rounded to three decimal places). The difference in  empirical standard deviation is also the biggest for Example $3$ ($0.242$), for Example $1$ is $0.068$ and for Example $2$ is only $0.034$. \\

\begin{figure}[h]
\centering
\subfloat[][Example 1]{
\includegraphics[width=0.45\linewidth]{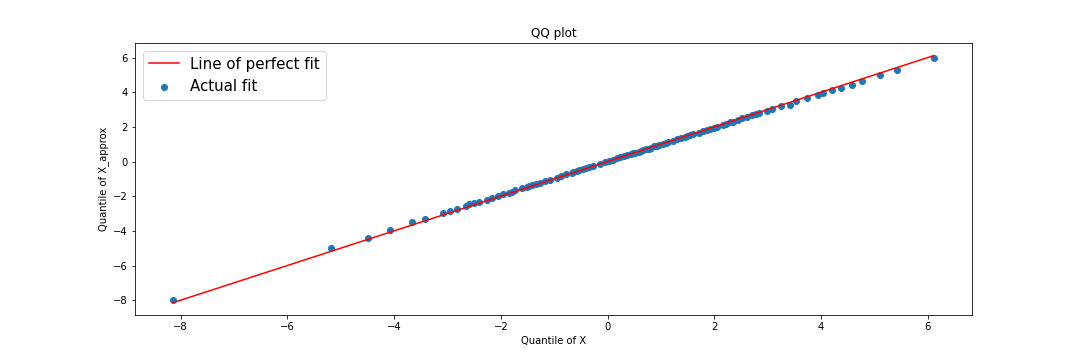}}
\subfloat[][Example 2]{
\includegraphics[width=0.45\linewidth]{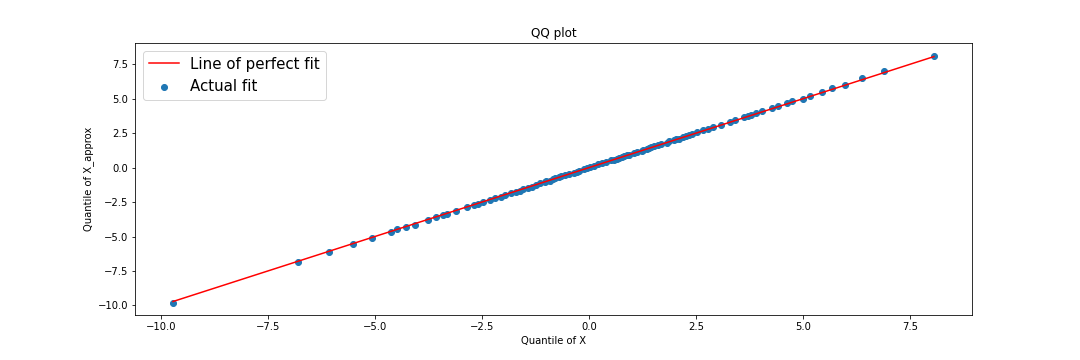}}
\quad
\subfloat[][Example 3]{
\includegraphics[width=0.45\linewidth]{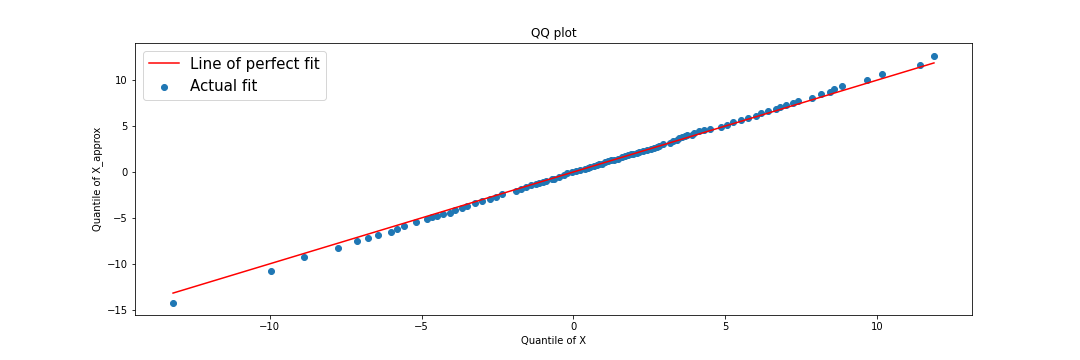}}
\quad
\subfloat[][Example 4]{
\includegraphics[width=0.45\linewidth]{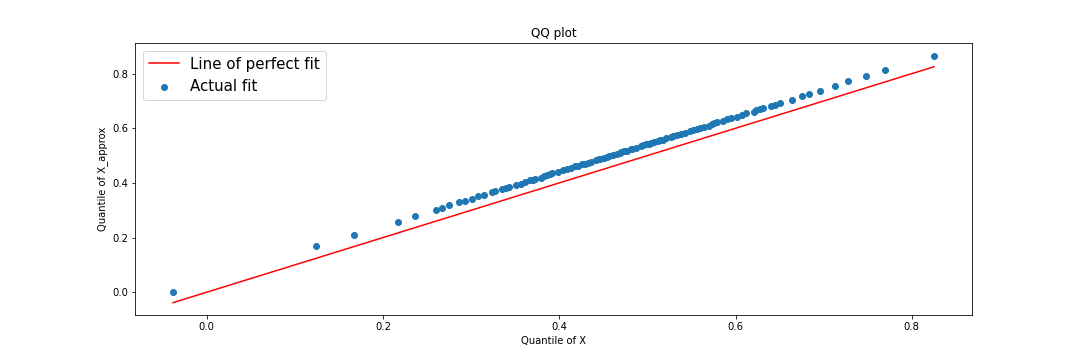}}

\caption{QQ plots - comparison of both distributions.}

\label{fig_qq-plots}
\end{figure}

 We also used quantile-quantile plot (QQ plot) method to compare both distributions at point $T$, presented in Figure \ref{fig_qq-plots}. Plots are consistent with previous observations. \\

We can conclude that empirical distribution of $\hat{X} = X(T, \hat{\sigma})$ is very similar to $X = X(T, \sigma)$ and for practical applications $X(T, \sigma)$ can be replaced by $X(T, \hat{\sigma})$.
\begin{table}
\begin{tabular}{ |c|c|c|c|c|c|c|c|c| }
 \hline
  & \multicolumn{2}{|c|}{Example 1} &
  \multicolumn{2}{|c|}{Example 2} & \multicolumn{2}{|c|}{Example 3} & \multicolumn{2}{|c|}{Example 4} \\
 \hline
     & X real & X approx & X real & X approx & X real & X approx & X real & X approx  \\
  \hline

empirical mean  & 0.515  & 0.515 & 0.495 & 0.497 & 1.243 &  1.256  & 0.477 &  0.519 \\
 \hline
empirical std dev & 2.456 & 2.388 & 3.168 & 3.202 & 4.835 &  5.077  & 0.144 &  0.143 \\
 \hline
\end{tabular} \caption{\label{histogram_values}Empirical distribution values for each example - mean and standard deviation} 
\end{table}
\begin{table}
\begin{tabular}{ |c|c|c|c|c| }
 \hline
& Example 1 & Example 2 & Example 3 & Example 4 \\
 \hline
KS statistics & 0.015 & 0.009 & 0.028 & 0.132 \\
 \hline
$p$-value of KS test & 0.99987 & 0.99999 & 0.82821 & 5.19e-08 \\
 \hline
\end{tabular} \caption{\label{distance_measures} Comparison of distributions metrics for all the examples} 
\end{table}

\subsubsection{\textbf{Statistical test}} 
We also used the Kolmogorov-Smirnov test with the null hypothesis that both the distributions $P$, $Q$ are identical, based on statistic $D$, which is given by
\begin{equation}
     D_{n,m} = \max_{x \in \mathbb{R}} |F_P(x) - F_Q(x)|,
\end{equation}
where $n$ - number of observations from distribution $P$ (Sample $1$), $m$ - number of observations from distribution $Q$ (Sample $2$). In our case $n = m = 1000$. $F_P$ is CDF of $P$ and $F_Q$ CDF of $Q$.

Numerical results are presented in Table \ref{distance_measures}. In Examples $1$ - $3$ the $p$-values is very high and there is no reason to reject null hypothesis. value is very high and there is no reason to reject null hypothesis. In Example $4$, the $p$-value is small enough and we should consider null hypothesis to be false, so we conclude that the distributions are not exactly the same. However, looking at the results of the empirical mean and variance, we can conclude that the approximation in real applications is very good.

\subsubsection{\textbf{Theorems 1 and 2 - empirical verification}} 
We also empirically validated the inequalities in  Theorems \ref{dist_X1X2}, \ref{dist_X1X2_discont}. The distance
\begin{equation*}
    \Bigl(\mathbb{E}|X_1(T)-X_2(T)|^2\Bigr)^{1/2}
\end{equation*}
was approximated by 
\begin{equation*}
    L_{empirical} = \sqrt{\frac{1}{N} \sum_{i=1}^{N} \Big| X_{(i)}(T) - \hat{X}_{(i)}(T) \Big| ^2},
\end{equation*}
where the notation was introduced in section \ref{desc_stat}. 

For the expression
\begin{equation*}
\|\Theta_1-\Theta_2\|^{\alpha_1}_{\infty}+\|\Theta_1-\Theta_2\|^{\alpha_2}_{\infty},
\end{equation*}
we took $\alpha_1 = \alpha_2 = 1$ and computed as
\begin{equation*}
R_{empirical}  = 2 \cdot \max\limits_{1 \leq i \leq n} \Big| \Big( \hat{\sigma} - \sigma \Big) ^{(i)} \Big| ,
\end{equation*}
where $n$ is number of points in mesh.

Results for our three examples are presented in Table \ref{theorem1_values}. $C_{empirical}$ was simply computed as $\frac{L_{empirical}}{R_{empirical}}$.

\begin{table}
\begin{tabular}{ |c|c|c|c|c| }
 \hline
& Example 1 & Example 2 & Example 3 & Example 4 \\
 \hline
$L_{empirical}$ & 0.081 & 0.060 & 0.449 & 0.041 \\
 \hline
$R_{empirical}$  & 0.597 & 0.503 & 1.252 & 2.223 \\
 \hline
$C_{empirical}$ & 0.136 & 0.119 & 0.359 & 0.019 \\
 \hline
\end{tabular} \caption{\label{theorem1_values} Empirical values for components of Theorem \ref{dist_X1X2} \& \ref{dist_X1X2_discont} inequality.} 
\end{table}
\subsubsection{\textbf{Summary and discussion}} 
Conducted experiments clearly show that our approximation method has real potential to manage time-dependent parameter estimation. Proposed approach possesses a significant limitation, it assumes the constant sign for the diffusion coefficient. When the assumption gets violated, it introduces the ambiguity in the approximation, i.e., positive and negative values of $b$ become indistinguishable. The ability to estimate the time-dependent parameters of coefficients provides the novelty of this paper. Nevertheless, difficulties emerge when estimating parameters associated with drift, even though the loss function incorporates parameters from both drift and diffusion. Recognizing the need to simultaneously tackle the estimation of drift parameters and time-dependent parameters in diffusion presents an opportunity for future research. Currently, known approaches are to simplify the time dependency and use piecewise-constant functions. The crucial part is formulating the loss function based on the maximum likelihood approach, which gets then translated into an optimization problem. Finally, for the optimization problem, we use deep learning techniques and employ TensorFlow - an efficient deep learning framework.
\section{Conclusions and future work}\label{section5}
Our results demonstrate the effectiveness of our proposed approach in accurately estimating time-dependent coefficients in dynamic systems. The combination of deep learning techniques and the maximum likelihood approach provides a powerful and versatile tool for parameter estimation. The ability to obtain reliable parameter estimates from a single trajectory makes our method particularly useful in scenarios where data collection is limited or expensive. Overall, our work contributes to the growing body of research on parameter estimation in SDE-based models and provides a valuable tool for researchers and practitioners in many fields. 
 As future work, we intend to explore the application of our approach to more complex systems and investigate its potential in real-time parameter estimation for dynamic systems control and optimization.

\appendix
\section{}\label{appendixA}

The complete set of codes for our project can be found on our GitHub repository: \\
\href{https://github.com/mwiacek8/DL-based-estimation}{REPOSITORY}. The following section outlines essential components of the code, focusing on key aspects related to regression and SDEs-based models.
%
\subsection{Regression}
In the Listings \ref{list1}--\ref{list3} we present how to simulate this type of data and in the Figures \ref{fig_reg_case1} - \ref{fig_reg_case3} we present our results for $n=3000$ number of points in each dimension. The function $\texttt{create\_sin\_data}$ takes the number of points as a parameter.

\begin{lstlisting}[label= list1, language=Python, caption=Generation of synthetic data - case 1]
# CASE 1 - variable mu_1, mu_2, constant sigma_1, sigma_2, rho

def create_sine_data_1(n=3000):
    np.random.seed(32)
    t = np.linspace(0, 2 * np.pi, n)

    x1 = 0.5 + np.sin(t)
    x2 = np.cos(t)

    mean_list = list(zip(x1, x2))

    sigma_1 = .1
    sigma_2 = .15
    rho = 0.5

    new_x1 = []
    new_x2 = []

    for zz1, zz2 in zip(x1, x2):
        cov = [[np.square(sigma_1), rho * sigma_1 * sigma_2], [rho * sigma_1 * sigma_2, np.square(sigma_2)]]

        rand_multi = np.random.multivariate_normal((0, 0), cov, 1)

        new_x1.append(zz1 + rand_multi[0][0])
        new_x2.append(zz2 + rand_multi[0][1])

    x1 = np.concatenate((np.ones(int(n / 10)) * 0.5, new_x1, np.ones(int(n / 10)) * 0.5))
    x2 = np.concatenate((np.ones(int(n / 10)), new_x2, np.ones(int(n / 10))))

    t_ = np.concatenate((np.linspace(-1, 0, int(n / 10)), t, np.linspace(max(t), max(t) + 1, int(n / 10))))

    return t_, x1, x2
\end{lstlisting}

\begin{lstlisting}[label= list2, language=Python, caption=Generation of synthetic data - case 2]
# CASE 2 - variable mu_1, mu_2, sigma_1, sigma_2, constant rho

def create_sine_data_2(n=3000):
    np.random.seed(32)
    t = np.linspace(0, 2 * np.pi, n)

    x1 = 0.5 + np.sin(t)
    x2 = np.cos(t)

    sigma_1 = .1
    sigma_2 = .15
    rho = 0.5

    new_x1 = []
    new_x2 = []

    for zz1, zz2 in zip(x1, x2):
        cov = [[np.square(sigma_1 * np.abs(zz1)), rho * sigma_1 * sigma_2 * np.abs(zz1) * np.abs(zz2)],
               [rho * sigma_1 * sigma_2 * np.abs(zz1) * np.abs(zz2), np.square(sigma_2 * np.abs(zz2))]]

        rand_multi = np.random.multivariate_normal((0, 0), cov, 1)

        new_x1.append(zz1 + rand_multi[0][0])
        new_x2.append(zz2 + rand_multi[0][1])

    x1 = np.concatenate((np.ones(int(n / 10)) * 0.5, new_x1, np.ones(int(n / 10)) * 0.5))
    x2 = np.concatenate((np.ones(int(n / 10)), new_x2, np.ones(int(n / 10))))

    t_ = np.concatenate((np.linspace(-1, 0, int(n / 10)), t, np.linspace(max(t), max(t) + 1, int(n / 10))))

    return t_, x1, x2
\end{lstlisting}

\begin{lstlisting}[label= list3, language=Python, caption=Generation of synthetic data - case 3]
# CASE 3 - variable mu_1, mu_2, sigma_1, sigma_2, rho

def create_sine_data_3(n=3000):
    np.random.seed(32)
    t = np.linspace(0, 2 * np.pi, n)

    x1 = 0.5 + np.sin(t)
    x2 = np.cos(t)

    sigma_1 = .1
    sigma_2 = .15
    rho = 0.5

    new_x1 = []
    new_x2 = []

    for zz1, zz2 in zip(x1, x2):
        cov = [[np.square(sigma_1) * np.abs(zz1), rho * sigma_1 * sigma_2 * np.abs(zz1) * np.abs(zz2)],
               [rho * sigma_1 * sigma_2 * np.abs(zz1) * np.abs(zz2), np.square(sigma_2) * np.abs(zz2)]]

        rand_multi = np.random.multivariate_normal((0, 0), cov, 1)

        new_x1.append(zz1 + rand_multi[0][0])
        new_x2.append(zz2 + rand_multi[0][1])

    x1 = np.concatenate((np.ones(int(n / 10)) * 0.5, new_x1, np.ones(int(n / 10)) * 0.5))
    x2 = np.concatenate((np.ones(int(n / 10)), new_x2, np.ones(int(n / 10))))

    t_ = np.concatenate((np.linspace(-1, 0, int(n / 10)), t, np.linspace(max(t), max(t) + 1, int(n / 10))))

    return t_, x1, x2
\end{lstlisting}

Our implementations of the loss functions are presented in Listings \ref{list4}--\ref{list7}. This loss function can be used in the neural network, so it compares the results obtained from the network with the training data set. The loss function is constructed so that the network is able to learn on how to recreate the parameters $\Theta$.

\begin{lstlisting}[label= list4,language=Python, caption=Loss function for two dimensional regression.]
def my_NLL_loss(y_true, y_pred):
    y_true_1 = tf.slice(y_true, [0, 0], [-1, 1])
    y_true_2 = tf.slice(y_true, [0, 1], [-1, 1])

    mu_1 = tf.slice(y_pred, [0, 0], [-1, 1])
    mu_2 = tf.slice(y_pred, [0, 1], [-1, 1])
    sigma_1 = tf.math.exp(tf.slice(y_pred, [0, 2], [-1, 1]))
    sigma_2 = tf.math.exp(tf.slice(y_pred, [0, 3], [-1, 1]))

    rho = 2 * tf.math.sigmoid(tf.slice(y_pred, [0, 4], [-1, 1])) - 1

    a = sigma_1 * sigma_2 * tf.sqrt(1 - tf.square(rho))

    b1 = tf.square((mu_1 - y_true_1) / sigma_1) + tf.square((mu_2 - y_true_2) / sigma_2) - 2. * rho * (
            (mu_1 - y_true_1) * (mu_2 - y_true_2) / (sigma_1 * sigma_2))
    b2 = 2 * (1 - tf.square(rho))
    b = b1 / b2

    loss = tf.reduce_sum(tf.math.log(a) + b, axis=0)

    return loss
\end{lstlisting}

\subsection{SDEs}

\begin{lstlisting}[label= list5,language=Python, caption=Loss function for Examples $1$-$3$]
def SDE_params_loss(y_true, y_pred):
    sigma = tf.slice(y_pred, [0, 0], [-1, 1])
    x_diff = tf.slice(y_true, [0, 2], [-1, 1])
    h = tf.slice(y_true, [0, 3], [-1, 1])
    x = tf.slice(y_true, [0, 1], [-1, 1])
    t = tf.slice(y_true, [0, 0], [-1, 1])

    A = tf.math.log(2 * math.pi * h * tf.square((b_(t, x, sigma))))
    B = tf.square(x_diff - a_(t, x) * h) / (2 * h * tf.square(b_(t, x, sigma)))
    loss = tf.reduce_sum(A / 2.0 + B, axis=0)

    return loss
\end{lstlisting}

\begin{lstlisting}[label= list6,language=Python, caption=Loss function for Example $4$]
def SDE_params_loss(y_true, y_pred):
    mu = tf.slice(y_pred, [0, 0], [-1, 1])  # A     
    sigma = tf.math.exp(tf.slice(y_pred, [0, 1], [-1, 1]))  # B  
    x_diff = tf.slice(y_true, [0, 2], [-1, 1])
    h = tf.slice(y_true, [0, 3], [-1, 1])
    x = tf.slice(y_true, [0, 1], [-1, 1])
    t = tf.slice(y_true, [0, 0], [-1, 1])

    A = tf.math.log(2 * math.pi * h * tf.square((b_(t, x, mu, sigma))))
    B = tf.square(x_diff - a_(t, x, mu, sigma) * h) / (h * tf.square(b_(t, x, mu, sigma)))
    loss = tf.reduce_sum(0.5 * (A + B), axis=0)
    return loss
\end{lstlisting}

\begin{lstlisting}[label= list7,language=Python, caption=Procedure generating trajectories and prediction intervals for SDEs.]
def trajectories_intervals(T, n, k, Wiener=None):
    """ Function for generate sample trajectory of processes X(t) and X_approx(t) with prediction intervals
        The parameters of equation (functions a, b are globally defined)

    Args:
        T (float): end of the interval, generate data for t \in [0,T]
        n (int): number of grid points for simulation (using Euler-Maruyama method)
        k (int): number of trajectories to generate
        Wiener (bool): table of Wiener process for reproducing results, None if it have to be newly generated 

    Returns:
        T - array of time points (grid)
        X_real - trajectory of process generated using given functions a,b 
        X_approx - trajectory generated using functions a,b with the parameters approximated (output from the model) and the same Wiener process as before
        X_approx_lower - lower prediction limit for X_approx
        X_approx_upper - upper prediction limit for X_approx

    """

    h = float(T / n)
    h_sqrt = np.sqrt(h)
    X_real = []
    X_approx = []

    X_approx_lower_ = []
    X_approx_upper_ = []

    T_ = []

    for j in range(k):

        X_real_temp = x_0
        X_approx_temp = x_0

        x_real = [X_real_temp]
        x_approx = [X_approx_temp]
        x_approx_lower = [X_approx_temp]
        x_approx_upper = [X_approx_temp]

        t_temp = 0.0
        t = [t_temp]

        for i in range(1, n + 1):

            if Wiener is not None:
                dW = Wiener[i - 1]

            else:
                dW = np.random.normal(0, h_sqrt)

            X_r = X_real_temp + a(t_temp, X_real_temp) * h + b_real(t_temp, X_real_temp) * dW
            x_real.append(X_r)

            X_a = X_a = X_approx_temp + a_(t_temp, X_approx_temp, mu_[i], sigma_[i]) * h + b_(t_temp, X_approx_temp,
                                                                                              mu_[i], sigma_[i]) * dW
            x_approx.append(X_a)

            X_a_lower = X_approx_temp + a_(t_temp, X_approx_temp, mu_[i], sigma_[i]) * h - 2 * np.abs(
                b_(t_temp, X_approx_temp, mu_[i], sigma_[i])) * np.sqrt(h)
            X_a_upper = X_approx_temp + a_(t_temp, X_approx_temp, mu_[i], sigma_[i]) * h + 2 * np.abs(
                b_(t_temp, X_approx_temp, mu_[i], sigma_[i])) * np.sqrt(h)
            x_approx_lower.append(X_a_lower)
            x_approx_upper.append(X_a_upper)

            t_temp = i * h
            t.append(t_temp)

            X_real_temp = X_r
            X_approx_temp = X_a

        X_real.append(x_real)
        X_approx.append(x_approx)
        X_approx_lower_.append(x_approx_lower)
        X_approx_upper_.append(x_approx_upper)

        T_.append(t)

        if (len(np.array(X_real).flatten()) != len(np.array(X_approx).flatten())):
            print("Trajectories have different length")

    return np.array(T_).flatten(), np.array(X_real).flatten(), np.array(X_approx).flatten(), np.array(
        X_approx_lower_).flatten(), np.array(X_approx_upper_).flatten()
\end{lstlisting}

\section{}\label{appendixB}
We have the following result on existence and uniqueness of strong solution to \eqref{main_equation_1}.
\begin{fact}
\label{prop_sde_1}
    Under the assumptions (A0)--(A4), (B1)--(B4), (T1) the SDE \eqref{main_equation_1} has unique strong solution $X=X(x_0,a,b,\Theta)$, such that for all $p\in [2,+\infty)$
    \begin{equation}
        \mathbb{E}\Bigl(\sup\limits_{0\leq t\leq T}\|X(t)\|^p\Bigr)<+\infty.
    \end{equation}
\end{fact}
\begin{proof}
    Let us denote by $\tilde a(t,x)=a(t,x,\Theta(t))$, $\tilde b(t,x)=b(t,x,\Theta(t))$. From the imposed assumptions we get that $\tilde a(\cdot,x):[0,T]\to\mathbb{R}^d$, $\tilde b(\cdot,x):[0,T]\to\mathbb{R}^{d\times m}$ are Borel measurable for every $x\in\mathbb{R}^d$, and $\tilde a(t,x):\mathbb{R}^d\to\mathbb{R}^d$ is continuous for every $t\in [0,T]$. Moreover
    \begin{equation}
        \langle x-y, \tilde a(t,x)-\tilde a(t,y)\rangle \leq H_+(1+\|\Theta\|_{\infty})\cdot \|x-y\|^2,
    \end{equation}
    for all $t\in [0,T]$, $x,y\in\mathbb{R}^d$, and
    \begin{eqnarray}
       && \|\tilde a(t,x)\|\leq \|a(t,x,\Theta(t))-a(t,x,0)\|+\|a(t,x,0)\|\notag\\
       &&\leq L(1+\|x\|^q)\cdot\|\Theta(t)\|^{\alpha_1}+\|a(t,x,0)\|,
    \end{eqnarray}
    which implies for all $\varrho\geq 0$ that
    \begin{eqnarray}
        &&\sup\limits_{0\leq t\leq T}\tilde a^{\#}_{\varrho}(t)=\sup_{(t,x)\in [0,T]\times B(0,\varrho)} \|\tilde a(t,x)\|\leq L(1+\varrho^q)\cdot\|\Theta\|_{\infty}^{\alpha_1}\notag\\
        &&\quad\quad+ \sup_{(t,x)\in [0,T]\times B(0,\varrho)} \|a(t,x,0)\| < + \infty.
    \end{eqnarray}
    We also have for all $t\in [0,T]$, $x,y\in\mathbb{R}^d$
    \begin{equation}
        \|\tilde b(t,x)-\tilde b(t,y)\|\leq L(1+\|\Theta\|_{\infty})\cdot \|x-y\|,
    \end{equation}
    and
    \begin{eqnarray}
        &&\|\tilde b(t,0)\|\leq \|b(t,0,\Theta(t))-b(t,0,0)\|+\|b(t,0,0)\|\notag\\
        &&\leq L\|\Theta(t)\|^{\alpha_2}+\|b(t,0,0)\|,
    \end{eqnarray}
    which gives
    \begin{equation}
        \sup\limits_{0\leq t\leq T}\|\tilde b(t,0)\|\leq L\|\Theta\|_{\infty}^{\alpha_2}+\sup\limits_{0\leq t\leq T}\|b(t,0,0)\|<+\infty.
    \end{equation}
    Finally,
    \begin{equation}
        \sup\limits_{0\leq t\leq T}\|\tilde a(t,0)\|\leq \sup\limits_{0\leq t\leq T}\tilde a^{\#}_{0}(t)<+\infty,
    \end{equation}
    and for all $p\in [2,+\infty)$ we have
    \begin{equation}
        \|x_0\|^p+\Bigl(\int\limits_0^T \|\tilde a(t,0)\|\rd t\Bigr)^p+\Bigl(\int\limits_0^T \|\tilde b(t,0)\|^2\rd t\Bigr)^{p/2}<+\infty.
    \end{equation}
    Hence, from Theorem 3.21., page 169 in \cite{PARRAS} we get the thesis.
\end{proof}
The following version of Gronwall's lemma  follows from Lemma 6.63.I., page 582 in \cite{PARRAS}.
\begin{lemma} (Gronwall's lemma with quadratic term)
\label{lem_gr_qt}
    Let $\varphi:[0,T]\to [0,+\infty)$ be a bounded Borel measurable function and let there exist $a,\alpha,\beta\in [0,+\infty)$ such that for all $t\in [0,T]$
    \begin{equation}
    \label{as_phi_1}
         \varphi^2(t)\leq a+2\alpha\int\limits_0^t\varphi(s)\rd s+2\beta\int\limits_0^t\varphi^2(s)\rd s.
    \end{equation}
    Then for all $t\in [0,T]$
    \begin{equation}
        \varphi(t)\leq \sqrt{a}\cdot e^{\beta t}+\alpha\cdot\frac{e^{\beta t}-1}{\beta}.
    \end{equation}
\end{lemma}
\begin{proof}
    For all $t\in [0,T]$ we define
    \begin{equation}
        x(t)=\Bigl(a+2\alpha\int\limits_0^t \varphi(s)\rd s+2\beta\int\limits_0^t\varphi^2(s)\rd s\Bigr)^{1/2},
    \end{equation}
    which is a non-negative, bounded, and absolutely continuous function. From \eqref{as_phi_1} we have for all $t\in [0,T]$ that
    \begin{equation}
        \varphi^2(t)\leq x^2(t)
    \end{equation}
     and  therefore for almost all $t\in [0,T]$
    \begin{equation}
        x'(t)\cdot x(t)= \alpha\varphi(t)+\beta\varphi^2(t)\leq \alpha x(t)+\beta x^2(t).
    \end{equation}
    Hence, from Lemma 6.63.I., page 582 in \cite{PARRAS} we get for all $t\in [0,T]$ that
    \begin{equation}
        \varphi(t)\leq x(t)\leq x(0) e^{\int\limits_0^t \beta \rd s}+\int\limits_0^t\alpha e^{\int\limits_s^t \beta \rd r}\rd s,
    \end{equation}
    which implies the thesis.
\end{proof}
We also recall the classical Gronwall's lemma, see, for example, Corollary 6.60 in  \cite[page 580]{PARRAS}.
\begin{lemma}
\label{gron_cont}
    Let $\alpha:[t_0,T]\to\mathbb{R}$ be an integrable function, and let $\beta:[t_0,T]\to [0,+\infty)$ be non-decreasing. If there exists $C\geq 0$ such that for all $t\in [t_0,T]$
    \begin{equation}
        \alpha(t)\leq \beta(t)+C\int\limits_{t_0}^t\alpha(s)ds,
    \end{equation}
    then
    \begin{equation}
        \alpha(t)\leq \beta(t)e^{C(t-t_0)}.
    \end{equation}
\end{lemma}
Below we give proofs of Theorems \ref{dist_X1X2}, \ref{dist_X1X2_discont}.\newline\newline
\noindent
{\bf Proof of Theorem \ref{dist_X1X2}.} 
    Let us denote by $\tilde a_i(t,x)=a(t,x,\Theta_i(t))$, $\tilde b_i(t,x)=b(t,x,\Theta_i(t))$, $i=1,2$. Then for all $t\in [0,T]$
    \begin{equation}
        Z(t)=X_1(t)-X_2(t)=\int\limits_0^t F(s)\rd s+\int\limits_0^t G(s)\rd W(s),
    \end{equation}
    where $F(s)=\tilde a_1(s,X_1(s))-\tilde a_2(s,X_2(s))$, $G(s)=\tilde b_1(s,X_1(s))-\tilde b_2(s,X_2(s))$. From the It\^o formula (see Corollary 2.18., page 90 in \cite{PARRAS}) we get that 
    \begin{equation}
        \|Z(t)\|^2=\int\limits_0^t \Bigl(2\langle Z(s), F(s)\rangle+\|G(s)\|^2\Bigr) \rd s+2\int\limits_0^t\langle Z(s),G(s)\rd W(s)\rangle,
    \end{equation}
    where
    \begin{equation}
        \int\limits_0^t\langle Z(s),G(s)\rd W(s)\rangle=\sum\limits_{k=1}^d\sum\limits_{j=1}^m\int\limits_0^t Z^k(s)\cdot G^{kj}(s)\rd W^j(s).
    \end{equation}
We now have
\begin{eqnarray}
    &&\int\limits_0^T \mathbb{E}\Bigl[\|Z(t)\|^2\cdot\|G(t)\|^2\Bigr]\rd t\leq 2TL^2(1+\|\Theta_1\|^2_{\infty})\cdot\mathbb{E}[\|Z\|^4_{\infty}]\notag\\
    &&\quad\quad\quad +4TL^2(\|\Theta_1\|_{\infty}+\|\Theta_2\|_{\infty})^{2\alpha_2}\cdot\Bigl(\mathbb{E}[\|Z\|^2_{\infty}]+\mathbb{E}[\|X_2\|^2_{\infty}\cdot\|Z\|^2_{\infty}]\Bigr)<+\infty,
\end{eqnarray}
since, by Fact \ref{prop_sde_1} and H\"older inequality,
\begin{equation}
    \mathbb{E}[\|X_2\|^2_{\infty}\cdot\|Z\|^2_{\infty}]\leq \Bigl(\mathbb{E}[\|X_2\|^4_{\infty}]\Bigr)^{1/2}\cdot\Bigl(\mathbb{E}[\|Z\|^4_{\infty}]\Bigr)^{1/2}<+\infty.
\end{equation}
Hence, for all $t\in [0,T]$
\begin{equation}
    \mathbb{E} \int\limits_0^t\langle Z(s),G(s)\rd W(s)\rangle=0,
\end{equation}
and thus
\begin{equation}
\label{exp_Z2}
\mathbb{E}\|Z(t)\|^2=2\mathbb{E}\int\limits_0^t \langle Z(s), F(s)\rangle\rd s+\int\limits_0^t\mathbb{E}\|G(s)\|^2 \rd s.
\end{equation}
For all $t\in [0,T]$
\begin{equation}
    \|G(t)\|\leq L(1+\|\Theta_1\|_{\infty}) \cdot \|Z(t)\|+L(1+\|X_2\|_{\infty})\cdot\|\Theta_1-\Theta_2\|^{\alpha_2}_{\infty},
\end{equation}
and hence
\begin{eqnarray}
\label{est_int_G}
&&\int\limits_0^t\mathbb{E}\|G(s)\|^2\rd s\leq 2L^2(1+\|\Theta_1\|_{\infty})^2\cdot\int\limits_0^t\mathbb{E}\|Z(s)\|^2\rd s\notag\\
&&\quad\quad+ 4L^2(1+\mathbb{E}[\|X_2\|^2_{\infty}])\cdot \|\Theta_1-\Theta_2\|^{2\alpha_2}_{\infty}.
\end{eqnarray}
Moreover, for all $t\in [0,T]$, by the assumptions (A2), (A4) and Cauchy-Schwarz inequality
\begin{eqnarray}
    &&\langle Z(t),F(t)\rangle = \langle X_1(t)-X_2(t), a(t,X_1(t),\Theta_1(t))-a(t,X_2(t),\Theta_2(t))\rangle\notag\\
    &&=\langle X_1(t)-X_2(t), a(t,X_1(t),\Theta_1(t))-a(t,X_2(t),\Theta_1(t))\rangle\notag\\
    &&\quad\quad+\langle X_1(t)-X_2(t), a(t,X_2(t),\Theta_1(t))-a(t,X_2(t),\Theta_2(t))\rangle\notag\\
    &&\leq H(1+\|\Theta_2(t)\|)\cdot \|X_1(t)-X_2(t)\|^2\notag\\
    &&\quad\quad+|\langle X_1(t)-X_2(t), a(t,X_2(t),\Theta_1(t))-a(t,X_2(t),\Theta_2(t))\rangle|\notag\\
    &&\leq H_+(1+\|\Theta_2\|_{\infty})\cdot \|Z(t)\|^2+\|Z(t)\|\cdot \|a(t,X_2(t),\Theta_1(t))-a(t,X_2(t),\Theta_2(t))\|\notag\\
    &&\leq H_+(1+\|\Theta_2\|_{\infty})\cdot \|Z(t)\|^2+L(1+\|X_2\|^q_{\infty})\cdot\|\Theta_1-\Theta_2\|^{\alpha_1}_{\infty}\cdot\|Z(t)\|.
\end{eqnarray}
Therefore, by the H\"older inequality we have for all $t\in [0,T]$
\begin{eqnarray}
\label{exp_ZF}
    &&\mathbb{E}\int\limits_0^t \langle Z(s),F(s)\rangle\rd s\leq H_+(1+\|\Theta_2\|_{\infty})\cdot \int\limits_0^t\mathbb{E}\|Z(s)\|^2\rd s\notag\\
    &&\quad\quad+L\|\Theta_1-\Theta_2\|^{\alpha_1}_{\infty}\cdot\int\limits_0^t \mathbb{E}\Bigl[\|Z(s)\|\cdot (1+\|X_2\|^q_{\infty})\Bigr]\rd s\notag\\
    &&\leq H_+(1+\|\Theta_2\|_{\infty})\cdot \int\limits_0^t\mathbb{E}\|Z(s)\|^2\rd s\notag\\
    &&\quad\quad+L\|\Theta_1-\Theta_2\|^{\alpha_1}_{\infty}\cdot\Bigl(\mathbb{E}(1+\|X_2\|^q_{\infty})^2\Bigl)^{1/2}\cdot\int\limits_0^t\Bigl(\mathbb{E }\|Z(s)\|^2\Bigr)^{1/2}\rd s.
\end{eqnarray}
Combining \eqref{exp_Z2}, \eqref{est_int_G}, and \eqref{exp_ZF} we get for all $t\in [0,T]$ that
\begin{eqnarray}
    \varphi^2(t)\leq a+2\alpha\int\limits_0^t\varphi(s)\rd s+2\beta\int\limits_0^t\varphi^2(s)\rd s,
\end{eqnarray}
where $\varphi(t)=\Bigl(\mathbb{E}\|Z(t)\|^2\Bigr)^{1/2}$ is bounded Borel measuarble function while
\begin{eqnarray}
    && a = 4TL^2(1+\mathbb{E}[\|X_2\|^2_{\infty}])\cdot \|\Theta_1-\Theta_2\|^{2\alpha_2}_{\infty}\\
    &&\alpha=L\|\Theta_1-\Theta_2\|^{\alpha_1}_{\infty}\cdot\Bigl(\mathbb{E}(1+\|X_2\|^q_{\infty})^2\Bigl)^{1/2}\\
    &&\beta= H_+(1+\|\Theta_2\|_{\infty})+L^2(1+\|\Theta_1\|_{\infty})^2.
\end{eqnarray}
Applying Lemma \ref{lem_gr_qt} from the Appendix we get the thesis.\ \ \ $\square$
\newline\newline
\noindent
{\bf Proof of Theorem \ref{dist_X1X2_discont}.}
    By the It\^o formula we obtain for all $t\in [0,T]$ that
    \begin{eqnarray}
         &&|X_1(t)-X_2(t)|^2=2\int\limits_0^t(X_1(s))-X_2(s))(a(s,X_1(s))-a(s,X_2(s)))\rd s\notag\\
         &&\quad\quad+2\int\limits_0^t(X_1(s))-X_2(s))(\Theta_1(s)-\Theta_2(s))\rd W(s)\notag\\
         &&\quad\quad+\int\limits_0^t|\Theta_1(s)-\Theta_2(s)|^2 \rd s.
    \end{eqnarray}
By \eqref{X_Sol_zv_est} we have that
\begin{eqnarray}
    &&\mathbb{E}\int\limits_0^T|X_1(t)-X_2(t)|^2\cdot|\Theta_1(t)-\Theta_2(t)|^2\rd t\notag\\
    &&\leq 4T(\|\Theta_1\|_{\infty}^2+\|\Theta_2\|_{\infty}^2)\cdot\Bigl[\mathbb{E}\Bigl(\sup\limits_{0\leq t\leq T}|X_1(t)|^2\Bigr)+\mathbb{E}\Bigl(\sup\limits_{0\leq t\leq T}|X_2(t)|^2\Bigr)\Bigr]<+\infty.
\end{eqnarray}
Therefore, 
\begin{equation}
    \mathbb{E}\int\limits_0^t(X_1(s))-X_2(s))(\Theta_1(s)-\Theta_2(s))\rd W(s)=0,
\end{equation}
and hence
\begin{eqnarray}
    &&\mathbb{E}|X_1(t)-X_2(t)|^2=2\mathbb{E}\int\limits_0^t(X_1(s))-X_2(s))(a(s,X_1(s))-a(s,X_2(s)))\rd s\notag\\
    &&\quad\quad+\int\limits_0^t|\Theta_1(s)-\Theta_2(s)|^2 \rd s\notag\\
    &&\leq 2H_+\int\limits_0^t \mathbb{E}|X_1(s)-X_2(s)|^2\rd s+\int\limits_0^t|\Theta_1(s)-\Theta_2(s)|^2 \rd s.
\end{eqnarray}
Applying Lemma \ref{gron_cont} we get the thesis. \ \ \ $\square$
\newline\newline
\section{Monte Carlo forecasting for SDEs}\label{appendixC}
The procedure of Monte Carlo forecasting for SDEs with estimated parameter function $\Theta(\cdot,w^*)$ is defined implemented as follows:
\begin{itemize}
	\item [\bf 1)] draw independent random samples $Z_0,Z_1,\ldots,Z_{N-1}$ form the distribution $\mathcal{N}(0,1)$,
		\item [\bf 2)] let $\tilde x_0:=x_n$ and for $k=0,1,\ldots,N-1$ compute
		\begin{equation}
			\tilde x_{k+1}=\tilde x_{k}+a(t_k,\tilde x_k,{\Theta(t_k,w^*)})\cdot h+				\sqrt{h}\cdot b(t_k,\tilde x_k,{\Theta(t_k,w^*)})\cdot Z_k,
			\end{equation}		 
		\item [\bf 3)] having $\{\tilde x_k\}_{k=0,1,\ldots,N}$, take $\bar x_0:=x_n$ and compute for $k=1,2,\ldots, N$ the predictions $\{\bar x_k\}_{k=1,2,\ldots,N}$ and the prediction intervals $\{\mathcal{\tilde P}_k(\alpha)\}_{k=1,2,\ldots,N}$ as follows
		\begin{eqnarray}
			&&\bar x_k = \tilde\mu_k=\tilde x_{k-1}+a(t_{k-1},\tilde x_{k-1},{\Theta(t_k,w^*)})\cdot h,\\
			&&\tilde\sigma_k=\sqrt{h}\cdot |b(t_{k-1},\tilde x_{k-1},{\Theta(t_k,w^*)})|,\\
			&&\mathcal{\tilde P}_k(\alpha)=[\tilde\mu_k-q_{\alpha}\cdot\tilde\sigma_k, \tilde\mu_k+q_{\alpha}\cdot\tilde\sigma_k].
		\end{eqnarray}			
\end{itemize}
%
%

{\noindent\bf Acknowledgments}
\\
This research was supported by the National Center for Research and Development under contract POIR.01.01.01-00-1924/20.

\section*{References}
\begingroup
\renewcommand{\section}[2]{}%
\bibliographystyle{siam}
\bibliography{bibliography}
\endgroup

\end{document}